\PassOptionsToPackage{table,dvipsnames}{xcolor}
\documentclass[sigconf]{acmart}
\AtBeginDocument{%
  }

\setcopyright{acmlicensed}
\copyrightyear{2018}
\acmYear{2018}
\acmDOI{XXXXXXX.XXXXXXX}

\usepackage{microtype}
\usepackage{graphicx}
\usepackage{pifont}
\usepackage{multirow} 
\usepackage{arydshln} 
\usepackage{makecell}
\usepackage{booktabs} 
\usepackage{subcaption}
\usepackage{graphicx} 
\usepackage{amsmath}
\usepackage{mathtools}
\usepackage{amsthm}
\usepackage{stackengine} 
\usepackage{enumitem}
\usepackage{fontawesome5} 
\usepackage{marvosym} 
\copyrightyear{2026}
\acmYear{2026}
\setcopyright{cc}
\setcctype{by}
\acmConference[MM '26]{Proceedings of the 34th ACM International Conference on Multimedia}{November 10--14, 2026}{Rio de Janeiro, Brazil}
\acmBooktitle{Proceedings of the 34th ACM International Conference on Multimedia (MM '26), November 10--14, 2026, Rio de Janeiro, Brazil}
\acmDOI{10.1145/3767308.3836111}
\acmISBN{979-8-4007-2213-4/2026/11}

\begin{document}

\title{Look Clearly Before Answering: Mitigating Hallucinations in LVLMs via Saliency-Driven Perceptual Realignment}

\makeatletter
\let\old@fnsymbol\@fnsymbol
\renewcommand{\@fnsymbol}[1]{%
  \ifnum#1=1
    \ensuremath{\dagger}%
  \else
    \old@fnsymbol{#1}%
  \fi}
\makeatother 


\author{Pengxu Chen}
 \affiliation{%
  \institution{Xidian University}
  \city{Xi'an}
  \country{China}}
\email{chenpx@stu.xidian.edu.cn}

\author{Yao Zhu}
\authornote{ Co-corresponding authors}
\affiliation{%
  \institution{Tsinghua University}
  \city{Beijing}
  \country{China}}
\email{ee_zhuy@zju.edu.cn}

\author{Guangming Zhu}
\affiliation{%
  \institution{Xidian University}
  \city{Xi'an}
  \country{China}}
\email{gmzhu@xidian.edu.cn}

\author{Jun Sheng}
\affiliation{%
  \institution{Shanghai Road Transport Development Center}
  \city{Shanghai}
  \country{China}}
\email{shengjun0325@126.com}

\author{Jincai Huang}
\affiliation{%
  \institution{Hunan Institute of \\ Advanced Technology}
  \city{Changsha}
  \country{China}}
\email{huangjincai@nudt.edu.cn}

\author{Xiangyang Ji}
\affiliation{%
  \institution{Tsinghua University}
  \city{Beijing}
  \country{China}}
\email{xyji@tsinghua.edu.cn}

\author{Liang Zhang}
\authornotemark[1]
\affiliation{%
  \institution{Xidian University}
  \city{Xi'an}
  \country{China}}
\email{liangzhang@xidian.edu.cn}


\begin{abstract}
Large vision-language models (LVLMs) have demonstrated remarkable capabilities in multimodal understanding. However, they remain prone to hallucinations, generating responses that are inconsistent with the visual evidence. 
Existing mitigation methods largely address language-prior bias or cross-modal imbalance, while progressive visual degradation across perception and memory remains underexplored. In this work, we propose Saliency-Driven Perceptual Realignment (SDPR), a training-free framework that mitigates the degradation of visual awareness throughout inference. Specifically, we first introduce saliency-driven attention redistribution to release attention hijacked by non-semantic sink tokens, thereby recovering critical visual evidence. Second, we identify spatial distortion in the KV cache and propose saliency-driven cache alignment to preserve query-relevant visual features during generation.
Finally, we introduce prior-constrained contrastive decoding to penalize unfaithful predictions induced by dominant language priors. Through holistic perceptual realignment across the generation process, SDPR robustly mitigates hallucinations. Extensive experiments across diverse LVLM architectures show that SDPR outperforms state-of-the-art methods on both hallucination and general-purpose benchmarks, requiring no additional training and incurring minimal runtime overhead. The code is available \href{https://github.com/PengSyuChen/SDPR}{\color{blue}{here}}.

\end{abstract}

\maketitle

\begin{figure}[h] %
 \centering       
\vspace{-13pt} 
\includegraphics[width=\columnwidth]{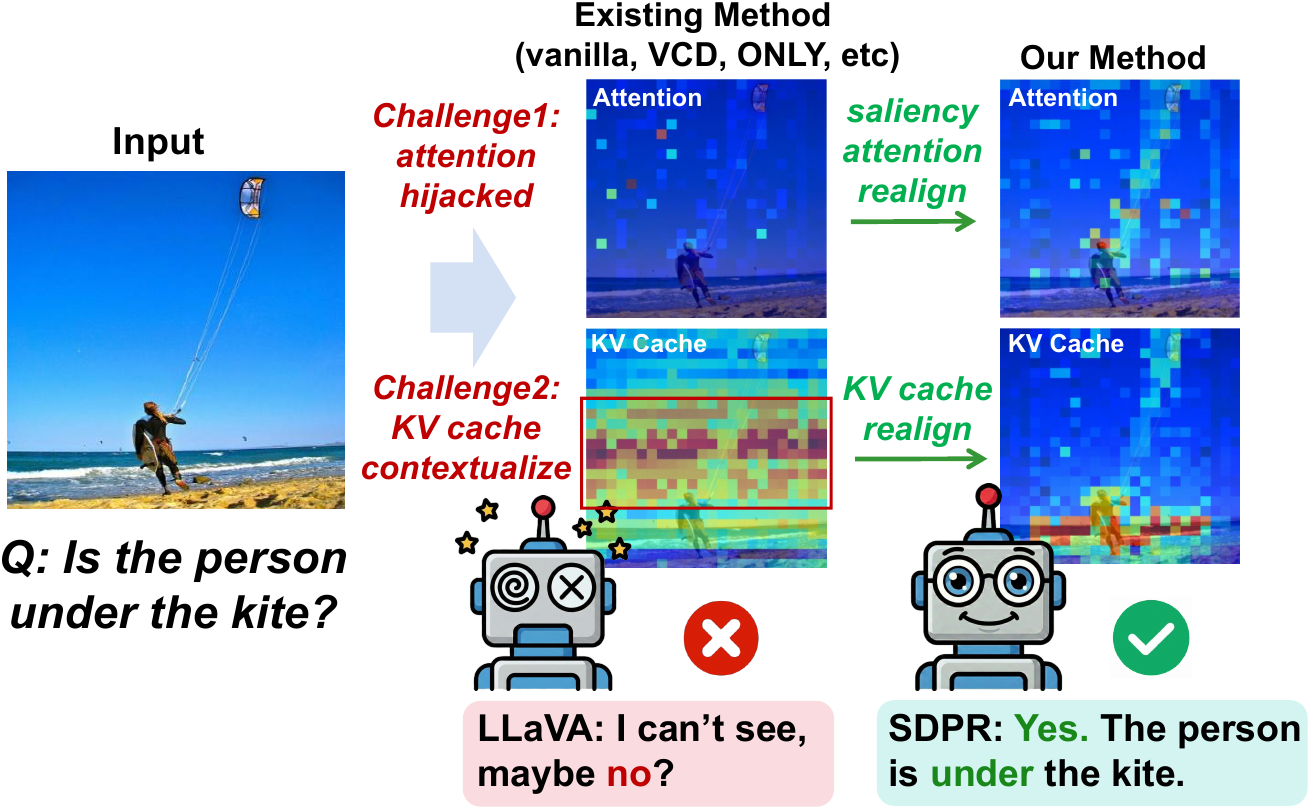} 
\vspace{-16pt}
\caption{While LVLMs struggle to capture and retain critical visual cues due to attention hijacking and KV-cache degradation, SDPR preserves visual alignment across perception and memory, leading to more reliable responses.}
 \label{firstFigure}
 \vspace{-15pt}
\end{figure}

\section{Introduction}
\label{introduction}
Recent advances in large vision-language models (LVLMs)
\cite{llava_1,qwen,internvl} have substantially improved multimodal
understanding, enabling these models to interpret complex visual scenes
and generate contextually coherent responses \cite{qwen,internvl}.
Despite this progress, LVLMs remain prone to hallucinations, generating
responses that are semantically plausible yet inconsistent with the visual
evidence. This limitation undermines their reliability and hinders deployment
in real-world applications that require factual accuracy.
\par

Prior studies have linked LVLM hallucinations to multiple factors, particularly
overreliance on language priors \cite{vcd,m3id} and imbalanced
cross-modal interactions \cite{only}. Recent studies identify a ``\emph{visual
amnesia}'' effect \cite{memvr}, in which the influence of question-relevant
visual evidence on token prediction gradually weakens as autoregressive
generation proceeds, while reliance on accumulated textual context increases.
We argue that these phenomena reflect a more fundamental limitation: the
progressive degradation of visual awareness across the perception and memory
stages. Specifically, non-semantic sink tokens hijack attention during
multimodal interaction, suppressing salient visual cues. Meanwhile, prefill-stage
causal processing \cite{qwen,internvl,transformer} asymmetrically contextualizes
visual tokens, inducing position-dependent distortion and weak query
conditioning in the KV cache. Together, these processes erode visual grounding,
shift next-token prediction toward language priors, and may even prompt guesses.
Without a \emph{clear look} at salient visual evidence before answering, LVLMs
are more likely to produce hallucinated responses.
\par

To further substantiate this hypothesis, we analyze the internal states of
LVLMs, as illustrated in Figure~\ref{firstFigure}. Our findings identify
distinct forms of visual degradation at the perception and memory stages,
which are quantitatively analyzed in Section~\ref{visual_degradation}.
First, the upper panel shows pronounced attention hijacking during multimodal
interaction in the language backbone: excessive attention is assigned to
non-semantic sink tokens associated with background regions, whereas salient
visual cues, such as the person and the kite string, receive little attention and thus exert less influence on subsequent token predictions.
Second, the PCA projection in the lower panel exhibits a sequence-ordered
pattern among cached visual representations. Under causal attention, later
visual tokens can aggregate earlier information, whereas earlier tokens cannot
access subsequent visual or question tokens. This asymmetric contextualization
induces sequence-dependent distortion and weakens query-conditioned retrieval
from the KV cache, making relevant visual evidence harder to access during
generation. In the illustrated case, this degradation shifts next-token
prediction toward language priors and leads to the hallucinated response shown
in Figure~\ref{firstFigure}. Together, these findings motivate a holistic
intervention strategy that recovers salient visual evidence during multimodal
interaction, preserves query-relevant visual information in the KV cache, and
constrains language-prior-driven predictions during decoding.
\par

Guided by these complementary objectives, we introduce Saliency-Driven
Perceptual Realignment (SDPR), a \emph{training-free} framework that
restores visual awareness throughout LVLM inference. First, to
rectify attention hijacking during multimodal interaction, we propose
Saliency-Driven Attention Redistribution (SDAR), which redirects attention
from non-semantic sink tokens toward salient visual features. Second, we
introduce Saliency-Driven Cache Alignment (SDCA) to reinforce the query
relevance of cached visual representations by injecting query-related
saliency maps into the key cache, thereby improving the retrieval of salient
visual evidence during autoregressive generation. Finally, we propose
Prior-Constrained Contrastive Decoding (PCD), which constructs a
language-prior reference from early hidden states and contrastively
constrains the decoding trajectory to suppress predictions dominated by
language priors rather than visual evidence
\cite{language_prior_cvpr25,language_prior_iclr26_1,
dola_language_prior_iclr24_1}.
As summarized in Figure~\ref{firstFigure}, SDPR jointly strengthens visual
grounding across perception, memory, and decoding, thereby suppressing
hallucinations and yielding visually grounded, factually faithful responses.

The contributions of this paper are as follows:
\begin{enumerate}[label=(\arabic*), leftmargin=*]
\item We systematically investigate the progressive degradation of visual
awareness throughout LVLM inference, revealing how deteriorating visual
grounding shifts generation toward language priors and induces hallucinated
responses.

\item We present SDPR, a \emph{training-free} framework that integrates
complementary interventions across perception, memory, and decoding to
restore visual awareness in LVLMs and mitigate multimodal hallucinations.

\item We conduct extensive experiments across diverse LVLMs. SDPR consistently
outperforms state-of-the-art methods on both hallucination and
general-purpose benchmarks, achieving up to a 6.93-point gain on POPE and
a 33.3\% reduction in CHAIR$_S$.
\end{enumerate}
\section{Related Work}
\label{related work}
\textbf{Large Vision-Language Models}. Building on the remarkable success
of LLMs, LVLMs have emerged as a leading paradigm for general-purpose
multimodal understanding. The LLaVA series \cite{llava_1,llava_1.5}
popularized a common architecture that connects a pre-trained vision encoder
to a powerful LLM backbone through a projection-based alignment module.
Subsequent models such as Qwen-VL \cite{qwen} and InternVL \cite{internvl}
have further improved multimodal performance through stronger visual
encoders, refined cross-modal alignment, and large-scale multimodal training.
Despite these advances, LVLMs remain prone to hallucinations and often
produce semantically plausible outputs that conflict with the provided
visual evidence. This limitation remains a major barrier to reliable
deployment in real-world scenarios. To overcome this limitation, we propose SDPR, a \emph{training-free}
inference-time framework that restores visual awareness and suppresses
multimodal hallucinations by counteracting degradation throughout inference.
\par
\textbf{Hallucination Mitigation in LVLMs.}
Early efforts primarily relied on
\textbf{\emph{training-centric methods}} such as supervised fine-tuning
and direct preference optimization
\cite{dpo_Hallucination_iccv1,dpo_hallucination_iclr1,
dpo_Hallucination_iccv2,
dpo_Hallucination_iccv3}. Although effective, these methods incur
substantial computational costs and require curated supervision or
preference data \cite{vcd}. To alleviate these burdens, recent work has
shifted toward training-free inference-time interventions.
\textbf{\emph{Decoding-based methods}} contrast logits from perturbed
counterparts to suppress hallucinated predictions \cite{vcd,m3id,only}.
However, due to their reliance on input-level perturbations
\cite{vcd,m3id}, these methods remain limited in addressing hallucinations
rooted in internal cross-modal interactions.
\textbf{\emph{Attention-intervention methods}} instead recalibrate internal
attention mechanisms within a single forward pass
\cite{vaf_clearsight,clvs,var,cai}, targeting intermediate cross-modal
fusion or visually sensitive attention heads to reinforce visual grounding
without repeated inference. Nevertheless, without explicit decoding-level constraints, such interventions
remain insufficient to suppress predictions dominated by strong language
priors \cite{only}. To address these complementary limitations, we propose SDPR, which integrates interventions across perception, memory, and prediction to restore visual awareness and mitigate hallucinations in LVLMs.
\begin{table}[t]
    \centering
    \caption{Attention concentration in hallucinatory and faithful responses.
    Lower values indicate weaker concentration.}
    \label{tab:sink_analysis}
    \vspace{-10pt}
    \resizebox{\linewidth}{!}{
    \begin{tabular}{l|cccc}
        \Xhline{0.8pt}
        & Gini $\downarrow$
        & $\mathrm{Attn}_{\mathrm{top}\text{-}1}\downarrow$
        & $\mathrm{Attn}_{\mathrm{top}\text{-}5}\downarrow$
        & $\mathrm{Attn}_{\mathrm{top}\text{-}10}\downarrow$ \\
        \hline
        Faithful
        & 2.85\%
        & 2.40\%
        & 10.79\%
        & 20.09\% \\
        Hallucinatory
        & 4.88\%
        & 4.29\%
        & 19.32\%
        & 35.82\% \\
        \emph{Gap Ratio}
        & \textcolor{red}{$\textbf{×1.71}$}
& \textcolor{red}{$\textbf{×1.79}$}
& \textcolor{red}{$\textbf{×1.79}$}
& \textcolor{red}{$\textbf{×1.78}$} \\
        \Xhline{0.8pt}
    \end{tabular}
    }
    \vspace{-10pt}
\end{table}
\section{Preliminary}
\textbf{Transformer-based LVLMs.} 
LVLMs employ a ViT-based encoder for perception and an LLM for reasoning.
Specifically, the visual input $I$ and query $Q$ are projected into a unified
latent space as tokens and then concatenated with the system-prompt tokens
$S$ to form the complete input sequence. During autoregressive inference,
the probability of predicting the next token $y_t \in \mathcal{V}$ is derived
from the final-layer hidden state $H_t^L$:
$y_t \sim p(y_t | I, Q, y_{<t}) = softmax(\phi(H^L_t))$
where $\phi(\cdot)$ denotes the language modeling head that projects the hidden state into vocabulary-wide logits.
\par
\textbf{LVLM Language Backbone.}
The language backbone comprises $L$ Transformer decoder layers.
Input embeddings $H^0$ are iteratively refined through Multi-Head
Self-Attention (MHSA) and Feed-Forward Networks (FFN) \cite{transformer}:
\begin{equation}
\hat{H}^{l}_{t}
=
\text{\emph{MHSA}}_l(H^{l}_{t}) + H^{l}_{t},
\quad
H^{l+1}_{t}
=
\text{\emph{FFN}}_l(\hat{H}^{l}_{t}) + \hat{H}^{l}_{t}.
\end{equation}
Each MHSA module contains $h$ parallel heads. For head $i$, the
query-to-sequence attention weights
$AW_i^l \in \mathbb{R}^{Q_{\mathrm{len}}\times N}$ are computed as
$AW_i^l=\operatorname{softmax}(Q_iK_i^T/\sqrt{d_k}+M)$, where $M$ is the
causal mask restricting each token to itself and preceding tokens, and
$Q_{\mathrm{len}}$ and $N$ are the query and sequence lengths. The head
outputs are concatenated and projected through $W_l^O$:
${MHSA}_l(H^l)={Concat}_{i=1}^{h}\left(AW_i^lV_i^l\right)W_l^O.
$
Finally, the next-token probability is obtained from $H_t^L$ through the
language modeling head.
\par

\textbf{KV Cache.}
To accelerate autoregressive inference, the KV cache stores preceding keys
and values at each decoder layer. During prefill, the model caches the keys
and values of all input tokens. At decoding step $t$, only the key $K_t^l$
and value $V_t^l$ of the newly generated token are computed and appended:
\begin{equation}
\mathcal{K}_t^l
=
\left[
\mathcal{K}_{t-1}^l;
K_t^l
\right],
\qquad
\mathcal{V}_t^l
=
\left[
\mathcal{V}_{t-1}^l;
V_t^l
\right].
\end{equation}

\section{Degradation of Visual Awareness in LVLMs}
\label{visual_degradation}
This section investigates the mechanisms underlying the degradation of visual
awareness in LVLMs, focusing on attention hijacking by non-semantic sink
tokens and asymmetric contextualization of cached visual representations during the 
perception and memory stages, respectively.
\subsection{Rethinking Visual Attention Sinks}
\begin{figure}[t]
 \centering       
\vspace{-6pt} 
\includegraphics[width=8.3cm]{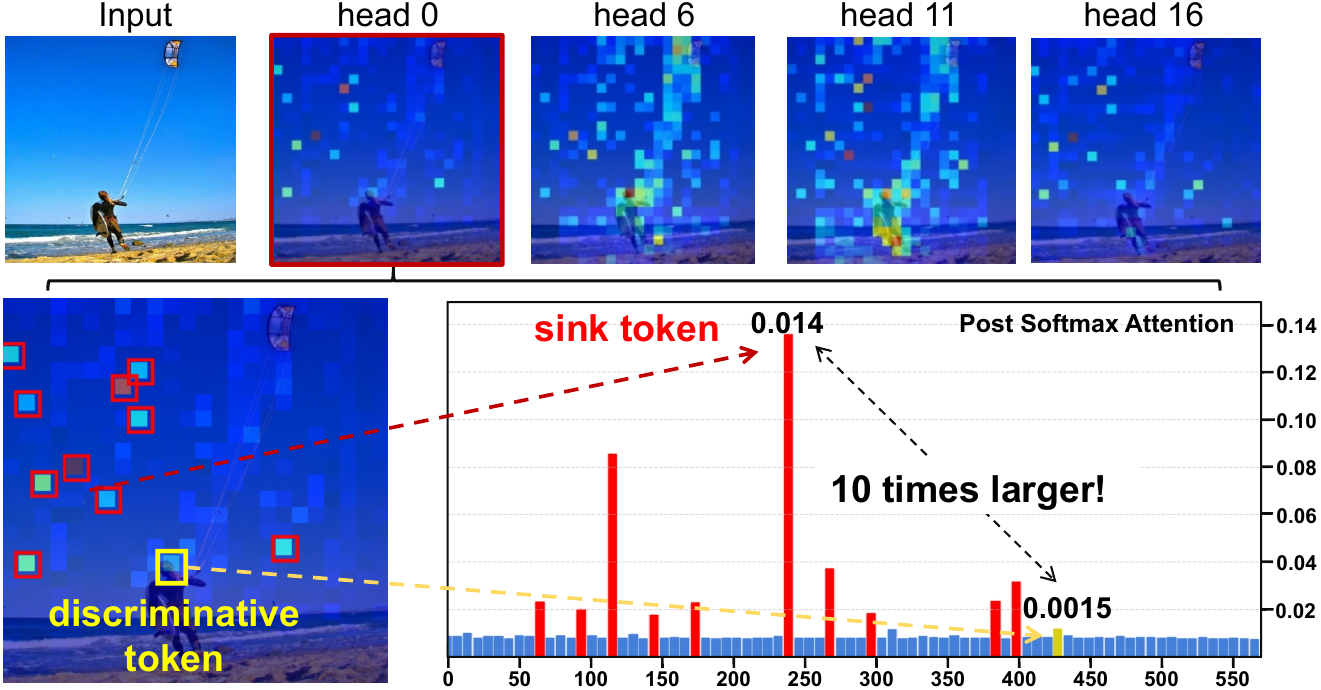} 
\vspace{-8pt}
\caption{Illustration of attention sink patterns. The upper panel shows LLaVA-1.5 attention heads; the lower panel details the attention distribution of a sink-dominated head.}
 \label{sink_figure_1}
 \vspace{-14pt}
\end{figure}

\textbf{Introduction.} Extensive research has highlighted the detrimental
effects of the attention-sink phenomenon, in which a few non-salient tokens
dominate the attention distribution and accumulate weights orders of
magnitude higher than those assigned to semantically meaningful tokens
\cite{gated_qwen_nips_best,sink_visual_iclr25,sink_iclr25_1}.
Consequently, the attention distribution becomes highly concentrated
(\emph{e.g.}, heads 0 and 16 in Figure~\ref{sink_figure_1}). In this work,
we investigate (1) whether attention-sink patterns contribute to
hallucinations, and (2) whether such sink states are permanent or reversible.
Ultimately, we explore the potential to redirect attention in sink-dominated
heads to mitigate hallucinations in LVLMs. \par 

\textbf{Observation 1: Rethinking Attention-Sink Patterns.}
To examine the relationship between attention-sink patterns and
hallucinations, we apply gradient-based saliency to MHSA attention weights
to estimate their contribution to model responses
\cite{HolisticCAM,CAV} on MME \cite{dataset_mme}. We further compare hallucinatory and faithful
responses using the Gini Score \cite{gini_nips25} and Top-$k$
Attention Ratio%
\footnote{(1) \emph{Gini Score}
($G=\frac{\sum_{i=1}^{n}\sum_{j=1}^{n}|x_i-x_j|}
{2n^2\bar{x}}$) measures global attention inequality; \\
(2) \emph{Top-$k$ Attention Ratio} measures the proportion of attention
mass assigned to the top $k$ tokens. Higher values indicate stronger
attention concentration.}.
As reported in Table~\ref{tab:sink_analysis}, hallucinatory responses
exhibit substantially higher attention concentration, reaching up to
1.79 times that of faithful responses. Together, these analyses reveal a
strong association between severe attention-sink patterns and LVLM
hallucinations.

We further examine the spatial distribution of attention sinks. As shown in
Figure~\ref{sink_figure_1}, we randomly sample several attention heads from
an intermediate layer of LLaVA-1.5, where cross-modal interactions are
frequent \cite{vaf_clearsight}. Two heads attend to salient content, whereas
the others exhibit pronounced attention-sink patterns. In the sink-dominated heads, all top-10 attended tokens correspond to
non-semantic background regions, while the discriminative token shown
in yellow receives only one-tenth as much attention as these sink tokens. This raises a key question: \emph{\textbf{Is the salient visual signal
irretrievably lost, or merely numerically overwhelmed?}}

\par
\begin{figure}[t]
    \centering
    \vspace{-7pt}
    \includegraphics[width=8.5cm]{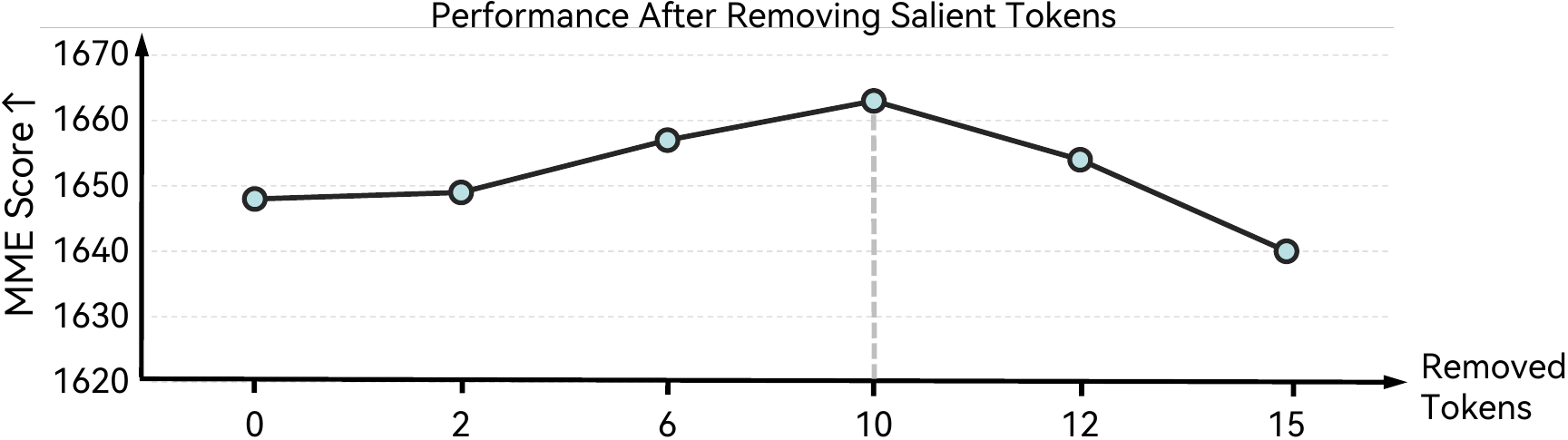}
    \vspace{-20pt}
    \caption{Numerical analysis of attention liberation. Removing 2--10
    top-attended tokens progressively improves overall performance, whereas
    removing more than 10 tokens leads to a clear decline.}
    \label{numerical_analysis_sink}
    \vspace{0pt}

    \centering
    \vspace{-1pt}
    \includegraphics[width=8.5cm]{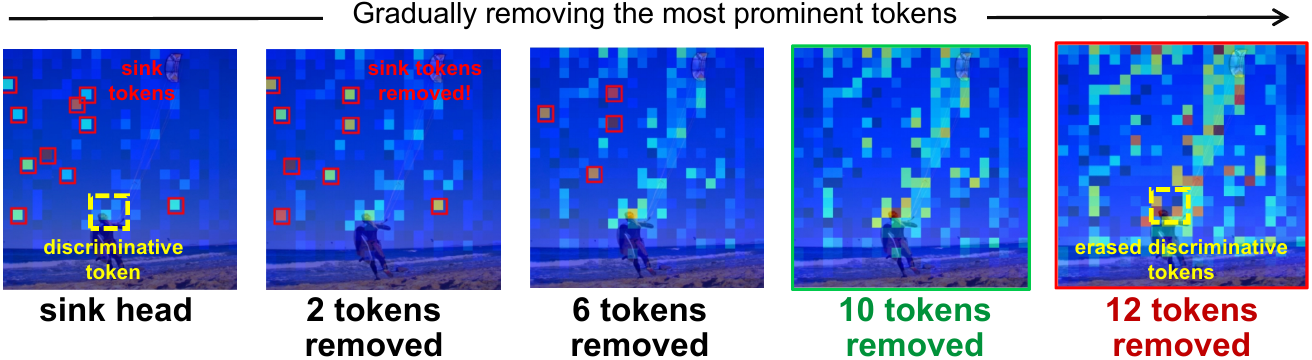}
    \vspace{-20pt}
    \caption{Progressive removal of the most significant tokens.}
    \label{sink_figure_2}
    \vspace{-15pt}
\end{figure}

\textbf{Motivation 1: Liberating Hijacked Attention.}
To answer this question, we conduct the progressive removal experiment
illustrated in Figure~\ref{sink_figure_2}. We iteratively remove the top-$k$
tokens with the highest attention weights and renormalize the remaining
distribution. Semantic features begin to emerge after removing the top
2--6 tokens, although background sink tokens remain dominant. Once the
top-10 tokens are removed, the previously suppressed signal
(\emph{i.e.}, the man's head) becomes dominant. This confirms that semantic information within sink-dominated heads is
preserved but numerically overwhelmed by dominant sink tokens. In other
words, \emph{\textbf{liberating hijacked attention can restore the
suppressed semantic responses of sink-dominated heads}}.
\par

\textbf{Motivation 2: The Dilemma of Naive Attention-Based Removal.}
Although the removal strategy effectively uncovers hidden semantics,
extending it further reveals a fundamental limitation. When $k$ increases
from 10 to 12, the newly recovered discriminative tokens, ranked 11th and
12th by attention weight, are also removed. This shows that a purely
magnitude-based removal criterion is \emph{semantic-agnostic}. Following a
``winner-takes-all'' principle, it removes high-attention tokens
indiscriminately and cannot distinguish meaningless attention sinks from
genuine semantic regions.
\par

The quantitative analysis in Figure~\ref{numerical_analysis_sink} further
confirms this pattern. As the number of removed sink tokens increases from 2 to 10, the MME score
steadily improves and then declines with further removal. This performance trend is consistent
with the case study in Figure~\ref{sink_figure_2} and exposes a fundamental
limitation of magnitude-based removal, which may suppress semantically
meaningful tokens once the removal threshold extends beyond dominant sink
tokens. This raises a critical question:
\textbf{\emph{Is there any solution that reconciles the elimination of the
sink effect with the preservation of semantic integrity in LVLM attention?}}

\subsection{Rethinking Visual Memory in the KV Cache}

\textbf{Introduction.}
The KV cache avoids redundant computation during autoregressive generation
by storing previously computed keys and values. However, causal attention
during prefill asymmetrically contextualizes visual tokens, introducing
position-dependent spatial distortion into their cached representations and
weakening query-relevant visual memory.

\begin{figure}[t!]
    \centering
    \vspace{-9pt}
    \includegraphics[width=8cm]{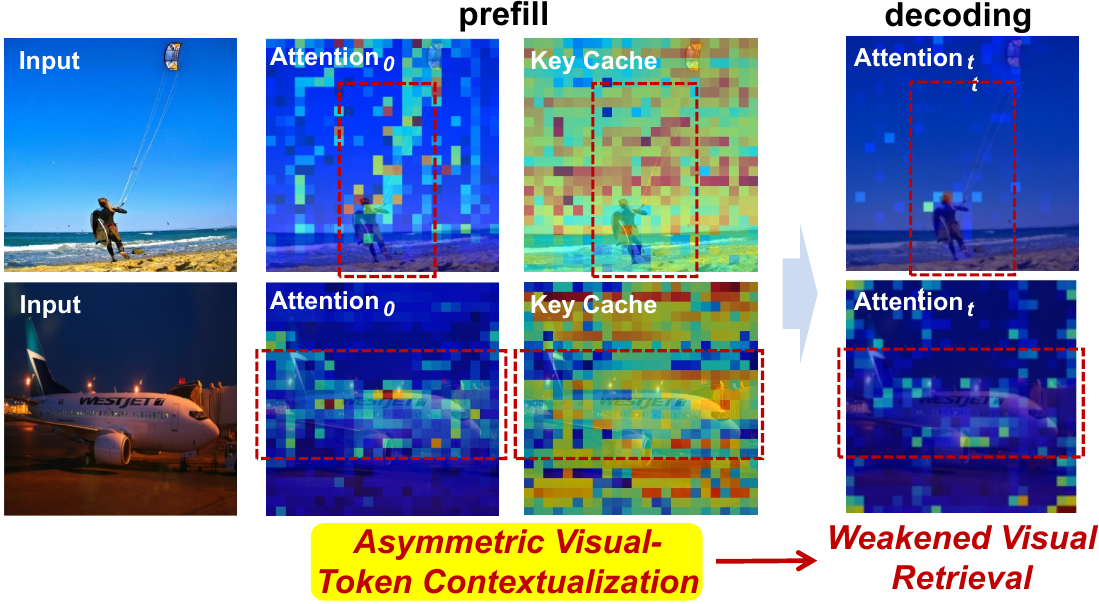}
    \vspace{-11pt}
    \caption{Spatial distortion in the key cache. Weakened retrieval of visual
evidence leads to diffuse decoding attention.}
    \label{kv_cache_analysis}
    \vspace{0pt}

    \centering
    \vspace{0pt}
    \includegraphics[width=8.5cm]{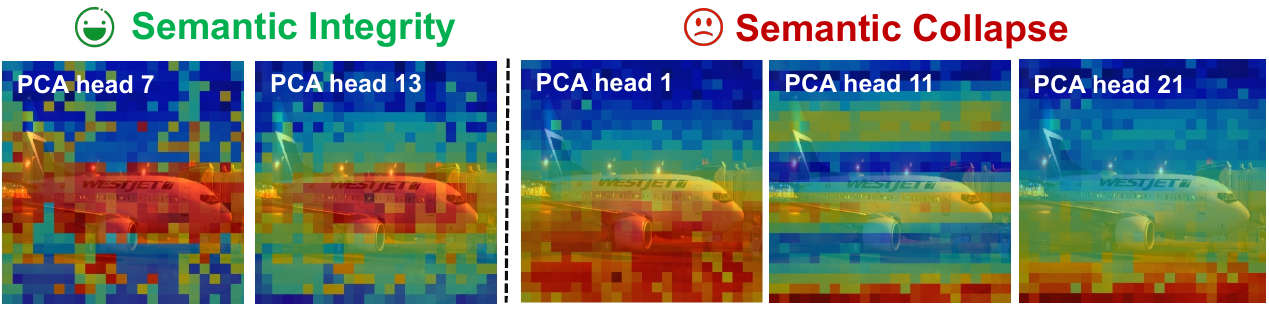}
    \vspace{-20pt}
    \caption{PCA of cached visual keys. Sequence-ordered patterns reveal
position-dependent distortion from asymmetric contextualization during prefill.}
    \label{kv_cache_pca}
    \vspace{-13pt}
\end{figure}

\textbf{Observation 2: Asymmetric Contextualization Distorts Cached Visual
Representations.}
Figure~\ref{kv_cache_analysis} reveals a discrepancy between the prefill and decoding stages. During prefill, attention
remains spatially coherent and focused on relevant regions, whereas
decoding-time attention is more diffuse over cached visual tokens,
indicating weakened retrieval of relevant visual evidence.
\par

Further analysis of the cached visual keys explains this retrieval degradation.
As shown in Figure~\ref{kv_cache_pca}, their PCA projections exhibit a
pronounced sequence-ordered pattern. Under causal attention, later visual
tokens can aggregate earlier visual context, whereas earlier tokens cannot
access subsequent visual tokens. Moreover, because the query follows the
visual sequence, cached visual keys cannot directly incorporate query
information during prefill. Since spatially arranged visual tokens are
contextualized through this one-way sequence, their cached representations
become increasingly dependent on sequence position, resulting in spatial
distortion and weak query conditioning. Consequently, relevant visual
evidence becomes harder to retrieve during decoding.
\par

\textbf{Motivation 3: Stronger Sequence-Position Dependence in Hallucinatory
Responses.}
If asymmetric contextualization contributes to hallucinations, hallucinatory
responses should exhibit stronger sequence-position dependence in their
cached visual keys. To test this hypothesis, we compute the absolute
Spearman rank correlation between the first principal-component scores of
cached visual keys and their visual-token sequence positions on the MME benchmark.\footnote{\emph{Spearman rank correlation} measures
the monotonic association between the first principal-component scores of
cached visual keys and their visual-token sequence positions. Larger
absolute values indicate stronger sequence-position dependence.} As shown in Table~\ref{tab:kv_analysis}, hallucinatory responses exhibit a
Spearman coefficient $1.62\times$ that of faithful responses, showing
substantially stronger sequence-position dependence. This result suggests
that the spatial distortion induced by asymmetric contextualization is more
pronounced in hallucinatory responses and is associated with weaker
query-conditioned visual retrieval. Therefore,
\emph{\textbf{counteracting position-dependent distortion in cached visual
representations is necessary to preserve query-relevant visual memory during
autoregressive generation}}.

\begin{table}[h!]
    \centering
    \vspace{-7pt}
    \caption{Quantitative analysis of cache spatial distortion.}
    \vspace{-9pt}
    \resizebox{0.4\linewidth}{!}{
        \begin{tabular}{l|c}
            \Xhline{1pt}
            & Spearman $\downarrow$ \\
            \hline
            Faithful & 0.0178 \\
            Hallucinatory & 0.0289 \\
            \emph{Gap Ratio} &
            \textcolor{red}{$\boldsymbol{\times 1.62}$} \\
            \hline
            \Xhline{0.8pt}
        \end{tabular}
    }
    \label{tab:kv_analysis}
    \vspace{-10pt}
\end{table}

\par

\section{Proposed Method}
\begin{figure*}[t!] 
 \centering 
 \vspace{-0.35cm} 
\includegraphics[width=\textwidth]{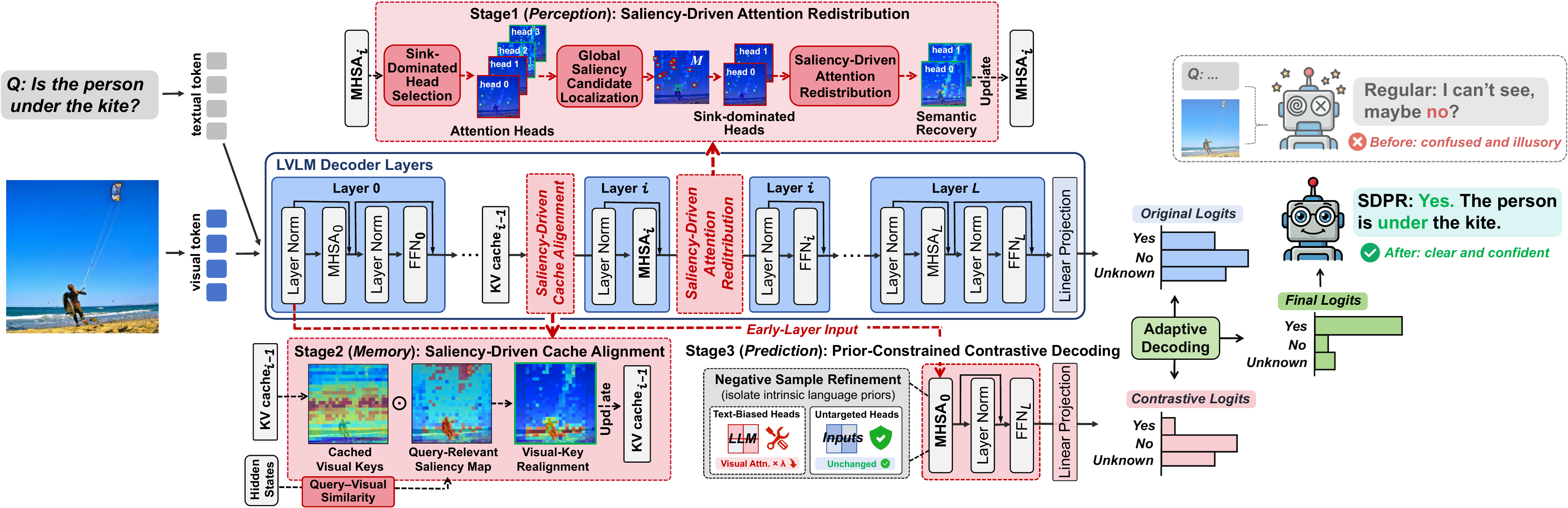}  
\vspace{-0.7cm} 
\caption{\textbf{Pipeline of SDPR.}
SDPR performs coordinated interventions across the perception, memory, and
prediction stages: (1) \emph{Perception}: SDAR redistributes
attention from non-semantic sink tokens toward salient visual features;
(2) \emph{Memory}: SDCA mitigates position-dependent distortion and
re-anchors query-relevant visual features in the key cache; and
(3) \emph{Prediction}: PCD contrasts the original logits with a
language-prior reference to suppress prior-dominated predictions.
}
 \vspace{-0.4cm} 
 \label{pipeline}
\end{figure*}

In response to the motivations outlined above, we detail the proposed method
in this section. As illustrated in Figure~\ref{pipeline}, SDPR consists of
three core components: \emph{Perception}: Saliency-Driven Attention
Redistribution to redirect hijacked attention within MHSA layers and restore
visual awareness; \emph{Memory}: Saliency-Driven Cache Alignment to mitigate
position-dependent distortion and re-anchor query-relevant visual features
in the key cache; and \emph{Prediction}: Prior-Constrained Contrastive
Decoding to suppress predictions dominated by language priors while
preserving visual grounding.

\subsection{Saliency-Driven Attention Redistribution}
Building on \emph{Motivations 1 and 2}, we propose Saliency-Driven
Attention Redistribution (SDAR) to attenuate non-semantic sink tokens
while recovering suppressed semantic evidence. SDAR uses a cross-head
saliency prior to localize dominant positions and leverages local spatial context
to guide head-specific redistribution. \par

\textbf{\emph{Sink-Dominated Head Selection}.}
Attention sinks yield sharply concentrated visual attention.
For head $h$ in the $l$-th MHSA layer, we compute the entropy of its
visual attention distribution $AW_h^{v,l}$:
\begin{equation}
    E_h^l =
    -\sum_{i=1}^{N_v}
    AW_{h,i}^{v,l}\log AW_{h,i}^{v,l}.
\end{equation}
Lower entropy indicates stronger concentration. We retain the $k$
lowest-entropy heads as putative sink-dominated heads,
$\mathcal{H}_{\mathrm{sink}}^l
=
\emph{Bottom}_{k}
(\{E_h^l\}_{h=1}^{H},\,k)$.
This criterion detects attention concentration but not semantic validity,
which is assessed using cross-head saliency and local spatial context.

\textbf{\emph{Global Saliency Candidate Localization}.}
To capture recurrent dominant responses while reducing head-specific
fluctuations, we aggregate the attention maps of the selected
low-entropy heads:
\begin{equation}
    M_i^l =
    \sum_{h\in\mathcal{H}_{\mathrm{sink}}^l}
    AW_{h,i}^{v,l}.
\end{equation}
To identify globally dominant positions shared across the selected heads,
we retain the top-$K$ positions in $M^l$ as saliency candidates.
Since these positions may correspond to either non-semantic sinks or
fine-grained semantic evidence, $M^l$ serves only to localize candidates
for subsequent head-specific refinement. \par
\begin{figure}[h!]
 \centering       
\vspace{3pt}
\includegraphics[width=8.5cm]{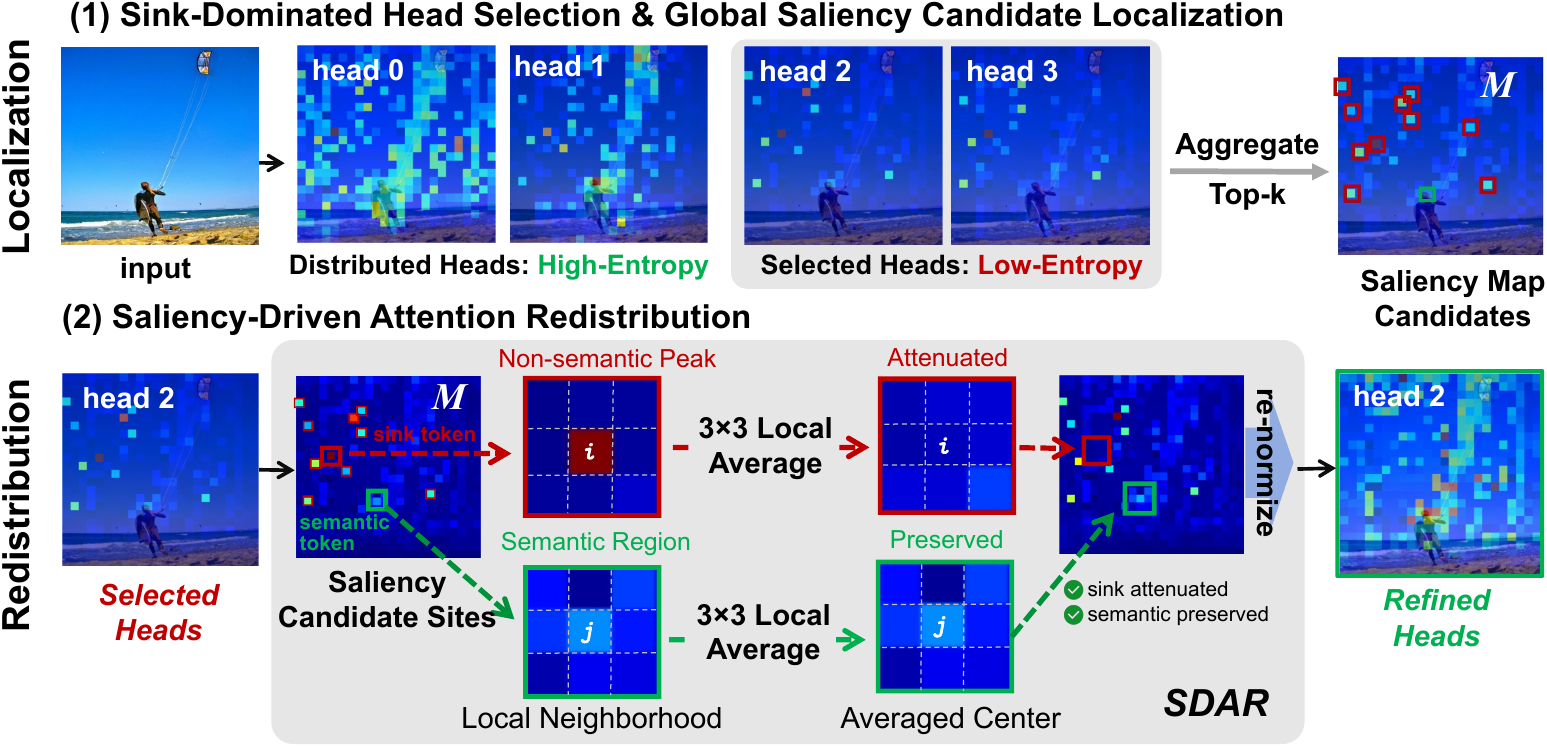} 
\vspace{-18pt}
\caption{\textbf{Saliency-Driven Attention Redistribution.}
(1) \emph{Localization}: low-entropy head aggregation locates global
candidates.
(2) \emph{Redistribution}: per-head averaging weakens non-semantic peaks and preserves semantic regions.}
 \label{method_1}
 \vspace{-15pt}
\end{figure} 

\textbf{\emph{Saliency-Guided Spatial Redistribution}.}
For each selected head, we refine the localized candidates according to
their spatial context. After mapping visual tokens back to the 2D grid,
we replace each candidate's attention score with the average over its
center-inclusive $3\times3$ neighborhood:
\begin{equation}
    \widehat{AW}_{h,i}^{v,l}
    =
    \frac{1}{|\mathcal{N}(i)|}
    \sum_{j\in\mathcal{N}(i)}
    AW_{h,j}^{v,l } ,\text{ where }i\in\mathcal{C}^l.
\end{equation}
Here, $\mathcal{C}^l$ denotes the top-$K$ candidate set derived from
$M^l$. This center-inclusive averaging weakens isolated non-semantic peaks by
mixing them with surrounding low responses, while retaining activations
within coherent semantic regions. We then apply \emph{min--max} scaling followed by $L_1$
normalization to restore a valid attention distribution. Consequently,
SDAR suppresses non-semantic sink responses while preserving
fine-grained semantic evidence.

\subsection{Saliency-Driven Cache Alignment}
Building on \emph{Motivation 3}, we introduce Saliency-Driven Cache
Alignment (SDCA) to mitigate position-dependent distortion and weak
query conditioning in visual keys. Since visual tokens precede
the query during prefill, their cached keys cannot encode subsequent
query information. SDCA therefore derives a query-relevant saliency
map and uses it to recalibrate visual-key retrieval.
\par
\textbf{\emph{Query-Relevant Saliency Estimation}.}
At a designated intermediate layer $l_s$, we extract the hidden states
$H^{l_s}\in\mathbb{R}^{N\times D}$ and partition them into query states
$H_q^{l_s}\in\mathbb{R}^{N_q\times D}$ and visual states
$H_v^{l_s}\in\mathbb{R}^{N_v\times D}$. We aggregate the query states
into a semantic anchor and compute its similarity to each visual token:
\begin{equation}
\begin{aligned}
    S_q
    =
    \sigma\left(
    \bar{h}_q \cdot (H_v^{l_s})^\top
    \right), \text{ where }\bar{h}_q
    =
    \frac{1}{N_q}
    \sum_{i=1}^{N_q} H_{q,i}^{l_s}.
\end{aligned}
\end{equation}
Here, $\sigma(\cdot)$ denotes \emph{min--max} normalization, and
$S_q\in\mathbb{R}^{N_v}$ measures the relevance of each visual token
to the current query. It provides explicit query conditioning for the
cached visual keys.

\textbf{\emph{Saliency-Guided Visual-Key Realignment}.}
Let $K_v^{l_s}\in\mathbb{R}^{N_v\times d_k}$ denote the cached visual
keys at layer $l_s$. We broadcast $S_q$ along the feature dimension and
reweight the visual keys as
\begin{equation}
    \widehat{K}_v^{l_s}
    =
    K_v^{l_s}\odot S_q.
\end{equation}
The aligned visual keys replace their original cache entries, while
non-visual keys and all cached values remain unchanged. Since keys
govern the matching between decoding queries and cached tokens, this
reweighting adjusts visual retrieval priority without directly modifying
the stored visual content. Consequently, SDCA reduces position-induced retrieval bias and facilitates
access to query-relevant visual evidence during generation.

\subsection{Prior-Constrained Contrastive Decoding}

Although SDAR and SDCA recover semantic visual evidence and improve its
retrieval from the cache, final predictions may still be dominated by
language priors. We therefore propose Prior-Constrained Contrastive
Decoding (PCD), which constructs a controlled language-prior reference
from early-layer features and adaptively contrasts it with the original
visually conditioned prediction.

\textbf{\emph{Language-Prior Reference Construction}.}
Prior studies suggest that task-relevant visual integration mainly
emerges in middle layers, whereas early representations are dominated by
pretrained linguistic patterns
\cite{dola_language_prior_iclr24_1,language_prior_cvpr25,
language_prior_iclr26_1}. We thus use the
normalized input $H_t^0$ to the first MHSA layer as a language-prior
representation before substantial multimodal fusion. To identify
text-biased heads, we define the modality preference disparity of head
$i$ as the ratio of visual to textual attention entropy:
\begin{equation}
    \mathcal{R}_{0,i}
    =
    \frac{E_{0,i}^{v}}{E_{0,i}^{t}}
    =
    \frac{
    -\sum_k p_{0,i,k}^{v}\log p_{0,i,k}^{v}
    }{
    -\sum_j p_{0,i,j}^{t}\log p_{0,i,j}^{t}
    }.
\end{equation}
Here, $p_{0,i,k}^{v}$ and $p_{0,i,j}^{t}$ denote normalized attention
probabilities over visual and textual tokens, respectively. A larger
$\mathcal{R}_{0,i}$ indicates diffuse visual attention but concentrated
textual attention, suggesting stronger textual bias. For heads exceeding
the layer-wise mean $\bar{\mathcal{R}}_0$, we selectively attenuate only
their visual attention:
\begin{equation}
    \widetilde{AW}_{0,i}^{v}
    =
    \begin{cases}
        \lambda AW_{0,i}^{v},
        & \mathcal{R}_{0,i}>\bar{\mathcal{R}}_0,\\
        AW_{0,i}^{v},
        & \text{otherwise},
    \end{cases}
\end{equation}
where $0<\lambda<1$; textual attention and untargeted heads remain unchanged.
This yields a constrained early representation
$\widetilde{H}_t^0$ that emphasizes language-prior tendencies. The original and reference logits are then obtained as
\begin{equation}
    f_t^{+}
    =
    \phi(H_t^L),
    \quad
    f_t^{-}
    =
    \phi\!\left(H_t^L+\widetilde{H}_t^0\right),
\end{equation}
where $\phi(\cdot)$ denotes the language modeling head. The residual
fusion retains the multimodal context in $H_t^L$ while injecting priors
captured by text-biased heads. Projecting both branches through the same
head makes the prior-enhanced reference logits $f_t^{-}$ directly comparable with
the original logits $f_t^{+}$.

\textbf{\emph{Adaptive Decoding}.}
During generation, we compute the $L_1$ distance between the original
and reference distributions over the candidate token set:
$
d_t
=
\sum_{y\in\mathcal{S}_t}
\left|
p_t^{+}(y)-p_t^{-}(y)
\right|.
$
A smaller $d_t$ indicates agreement between prior-based and visually
conditioned predictions, whereas a larger value suggests conflict between
language priors and visual evidence. Then, we select between the collaboration and contrast paths based on a threshold $\gamma$:
\begin{equation}
    f_t^{\mathrm{final}}
    =
    \begin{cases}
        f_t^{+}+\alpha_1 f_t^{-},
        & d_t<\gamma
        \text{ (collaboration)},\\
        (1+\alpha_2)f_t^{+}-\alpha_2f_t^{-},
        & d_t\geq\gamma
        \text{ (contrast)}.
    \end{cases}
\end{equation}
When the branches agree, the collaboration path reinforces shared
predictions. When they diverge, the contrast path subtracts the
language-prior reference to suppress prior-dominated candidates.

\section{Experiments}
In this section, we compare SDPR with existing hallucination-mitigation
methods through benchmark evaluations, ablations of the primary components,
and case studies. Additional results on general-purpose LVLM benchmarks,
hyperparameter studies, efficiency comparisons, and qualitative cases are
provided in the
\href{https://github.com/PengSyuChen/SDPR/blob/main/supplementary.pdf}
{supplementary material}.

\begin{table*}
\vspace{0pt}
    \centering
    \caption{
    \underline{POPE} performance comparison on different benchmarks using LLaVA-1.5-7B. $\uparrow$ indicates that higher values are better; the best results are shown in \textbf{bold}. Improvements over baseline are marked in green (\textbf{\color{ForestGreen}$\uparrow$}), while drops are in red (\textbf{\color{red}$\downarrow$}).
    }
    \vspace{-10pt}
    \resizebox{0.85\textwidth}{!}
{
\label{pope_llava}
\begin{tabular}{c|l|cccccccc} 
\Xhline{0.8pt}
\multicolumn{2}{c}{\multirow{2}{*}{\shortstack{\textbf{\emph{POPE}} \\ \textbf{\textit{LLaVA-1.5}}}}} & \multicolumn{2}{c}{Random} & \multicolumn{2}{c}{Popular} & \multicolumn{2}{c}{Adversarial} & \multicolumn{2}{c}{Average} \\ 
\cline{3-10} 
\multicolumn{2}{c}{} & Accuracy $\uparrow$    & F1-Score $\uparrow$     & Accuracy $\uparrow$  & F1-Score $\uparrow$ & Accuracy $\uparrow$  & F1-Score $\uparrow$ & Accuracy $\uparrow$  & F1-Score $\uparrow$ \\

\hline
\multirow{9}{*}{\rotatebox{90}{COCO}}    
& Vanilla  & 85.033 {\scriptsize\color{ForestGreen}$\uparrow$0.00} & 85.426 {\scriptsize\color{ForestGreen}$\uparrow$0.00} & 81.333 {\scriptsize\color{ForestGreen}$\uparrow$0.00} & 82.334 {\scriptsize\color{ForestGreen}$\uparrow$0.00} & 76.333 {\scriptsize\color{ForestGreen}$\uparrow$0.00} & 79.039 {\scriptsize\color{ForestGreen}$\uparrow$0.00} & 80.900 {\scriptsize\color{ForestGreen}$\uparrow$0.00} & 82.266 {\scriptsize\color{ForestGreen}$\uparrow$0.00} \\
& VCD~\emph{\scriptsize  CVPR'24} & 85.800~{\scriptsize\color{ForestGreen}$\uparrow$0.77} & 86.240~{\scriptsize\color{ForestGreen}$\uparrow$0.81} & 80.700~{\scriptsize\color{red}$\downarrow$0.63} & 81.945~{\scriptsize\color{red}$\downarrow$0.39} & 77.333~{\scriptsize\color{ForestGreen}$\uparrow$1.00} & 79.701~{\scriptsize\color{ForestGreen}$\uparrow$0.66} & 81.278~{\scriptsize\color{ForestGreen}$\uparrow$0.38} & 82.629~{\scriptsize\color{ForestGreen}$\uparrow$0.36} \\
& M3ID~\emph{\scriptsize  CVPR'24} & 86.733~{\scriptsize\color{ForestGreen}$\uparrow$1.70} & 87.010~{\scriptsize\color{ForestGreen}$\uparrow$1.58} & 82.300~{\scriptsize\color{ForestGreen}$\uparrow$0.97} & 83.222~{\scriptsize\color{ForestGreen}$\uparrow$0.89} & 77.700~{\scriptsize\color{ForestGreen}$\uparrow$1.37} & 79.940~{\scriptsize\color{ForestGreen}$\uparrow$0.90} & 82.244~{\scriptsize\color{ForestGreen}$\uparrow$1.34} & 83.391~{\scriptsize\color{ForestGreen}$\uparrow$1.13} \\
& VAR~\emph{\scriptsize  ICLR'25} & 84.700~{\scriptsize\color{red}$\downarrow$0.33} & 85.034~{\scriptsize\color{red}$\downarrow$0.39} & 81.400~{\scriptsize\color{ForestGreen}$\uparrow$0.07} & 82.375~{\scriptsize\color{ForestGreen}$\uparrow$0.04} & 75.800~{\scriptsize\color{red}$\downarrow$0.53} & 78.211~{\scriptsize\color{red}$\downarrow$0.83} & 80.633~{\scriptsize\color{red}$\downarrow$0.27} & 81.873~{\scriptsize\color{red}$\downarrow$0.39} \\
& VAF~\emph{\scriptsize  CVPR'25} & 83.333~{\scriptsize\color{red}$\downarrow$1.70} & 84.326~{\scriptsize\color{red}$\downarrow$1.10} & 85.566~{\scriptsize\color{ForestGreen}$\uparrow$4.23} & 85.964~{\scriptsize\color{ForestGreen}$\uparrow$3.63} & 74.366~{\scriptsize\color{red}$\downarrow$1.97} & 77.742~{\scriptsize\color{red}$\downarrow$1.30} & 81.088~{\scriptsize\color{ForestGreen}$\uparrow$0.19} & 82.677~{\scriptsize\color{ForestGreen}$\uparrow$0.41} \\
& ONLY~\emph{\scriptsize  ICCV'25} & 89.166~{\scriptsize\color{ForestGreen}$\uparrow$4.13} & 89.119~{\scriptsize\color{ForestGreen}$\uparrow$3.69} & 85.966~{\scriptsize\color{ForestGreen}$\uparrow$4.63} & 86.291~{\scriptsize\color{ForestGreen}$\uparrow$3.96} & 79.333~{\scriptsize\color{ForestGreen}$\uparrow$3.00} & 81.590~{\scriptsize\color{ForestGreen}$\uparrow$2.55} & 84.822~{\scriptsize\color{ForestGreen}$\uparrow$3.92} & 85.667~{\scriptsize\color{ForestGreen}$\uparrow$3.40} \\
& AvisC~\emph{\scriptsize  ACL'25} & 82.307~{\scriptsize\color{red}$\downarrow$2.73} & 83.502~{\scriptsize\color{red}$\downarrow$1.92} & 78.242~{\scriptsize\color{red}$\downarrow$3.09} & 80.534~{\scriptsize\color{red}$\downarrow$1.80} & 74.200~{\scriptsize\color{red}$\downarrow$2.13} & 77.706~{\scriptsize\color{red}$\downarrow$1.33} & 78.250~{\scriptsize\color{red}$\downarrow$2.65} & 80.581~{\scriptsize\color{red}$\downarrow$1.69} \\
& MoD~\emph{\scriptsize  ACL'25}  & 88.866~{\scriptsize\color{ForestGreen}$\uparrow$3.83} & 88.829~{\scriptsize\color{ForestGreen}$\uparrow$3.40} & 85.566~{\scriptsize\color{ForestGreen}$\uparrow$4.23} & 85.964~{\scriptsize\color{ForestGreen}$\uparrow$3.63} & 79.200~{\scriptsize\color{ForestGreen}$\uparrow$2.87} & 80.964~{\scriptsize\color{ForestGreen}$\uparrow$1.93} & 84.544~{\scriptsize\color{ForestGreen}$\uparrow$3.64} & 85.252~{\scriptsize\color{ForestGreen}$\uparrow$2.99} \\

\rowcolor{gray!20} 
\cellcolor{white}
& \textbf{SDPR~\emph{\scriptsize  ours}}  & \textbf{89.966}~\textbf{\scriptsize\color{ForestGreen}$\uparrow$4.93} & \textbf{89.674}~\textbf{\scriptsize\color{ForestGreen}$\uparrow$4.25} & \textbf{86.233}~\textbf{\scriptsize\color{ForestGreen}$\uparrow$4.90} & \textbf{86.347}~\textbf{\scriptsize\color{ForestGreen}$\uparrow$4.01} & \textbf{79.500}~\textbf{\scriptsize\color{ForestGreen}$\uparrow$3.17} & \textbf{81.817}~\textbf{\scriptsize\color{ForestGreen}$\uparrow$2.78} & \textbf{85.233}~\textbf{\scriptsize\color{ForestGreen}$\uparrow$4.33} & \textbf{85.946}~\textbf{\scriptsize\color{ForestGreen}$\uparrow$3.68} \\ 
\hline
\multirow{9}{*}{\rotatebox{90}{A-OKVQA}} 
& Vanilla & 82.133 {\scriptsize\color{ForestGreen}$\uparrow$0.00} & 83.874 {\scriptsize\color{ForestGreen}$\uparrow$0.00} & 75.400 {\scriptsize\color{ForestGreen}$\uparrow$0.00} & 79.069 {\scriptsize\color{ForestGreen}$\uparrow$0.00} & 67.000 {\scriptsize\color{ForestGreen}$\uparrow$0.00} & 73.656 {\scriptsize\color{ForestGreen}$\uparrow$0.00} & 74.844 {\scriptsize\color{ForestGreen}$\uparrow$0.00} & 78.866 {\scriptsize\color{ForestGreen}$\uparrow$0.00} \\
& VCD~\emph{\scriptsize  CVPR'24} & 81.666~{\scriptsize\color{red}$\downarrow$0.47} & 83.621~{\scriptsize\color{red}$\downarrow$0.25} & 74.900~{\scriptsize\color{red}$\downarrow$0.50} & 78.877~{\scriptsize\color{red}$\downarrow$0.19} & 67.000~{\scriptsize\color{ForestGreen}$\uparrow$0.00} & 73.809~{\scriptsize\color{ForestGreen}$\uparrow$0.15} & 74.522~{\scriptsize\color{red}$\downarrow$0.32} & 78.769~{\scriptsize\color{red}$\downarrow$0.10} \\
& M3ID~\emph{\scriptsize  CVPR'24} & 82.800~{\scriptsize\color{ForestGreen}$\uparrow$0.67} & 84.448~{\scriptsize\color{ForestGreen}$\uparrow$0.57} & 75.900~{\scriptsize\color{ForestGreen}$\uparrow$0.50} & 79.512~{\scriptsize\color{ForestGreen}$\uparrow$0.44} & 67.133~{\scriptsize\color{ForestGreen}$\uparrow$0.13} & 73.915~{\scriptsize\color{ForestGreen}$\uparrow$0.26} & 75.278~{\scriptsize\color{ForestGreen}$\uparrow$0.43} & 79.292~{\scriptsize\color{ForestGreen}$\uparrow$0.43} \\
& VAR~\emph{\scriptsize  ICLR'25} & 82.166~{\scriptsize\color{ForestGreen}$\uparrow$0.03} & 83.900~{\scriptsize\color{ForestGreen}$\uparrow$0.03} & 75.366~{\scriptsize\color{red}$\downarrow$0.03} & 79.047~{\scriptsize\color{red}$\downarrow$0.02} & 67.100~{\scriptsize\color{ForestGreen}$\uparrow$0.10} & 73.729~{\scriptsize\color{ForestGreen}$\uparrow$0.07} & 74.877~{\scriptsize\color{ForestGreen}$\uparrow$0.03} & 78.892~{\scriptsize\color{ForestGreen}$\uparrow$0.03} \\
& VAF~\emph{\scriptsize  CVPR'25} & 79.633~{\scriptsize\color{red}$\downarrow$2.50} & 82.435~{\scriptsize\color{red}$\downarrow$1.44} & 73.533~{\scriptsize\color{red}$\downarrow$1.87} & 78.091~{\scriptsize\color{red}$\downarrow$0.98} & 65.700~{\scriptsize\color{red}$\downarrow$1.30} & 73.210~{\scriptsize\color{red}$\downarrow$0.45} & 72.955~{\scriptsize\color{red}$\downarrow$1.89} & 77.912~{\scriptsize\color{red}$\downarrow$0.95} \\
& ONLY~\emph{\scriptsize  ICCV'25} & 86.066~{\scriptsize\color{ForestGreen}$\uparrow$3.93} & 87.154~{\scriptsize\color{ForestGreen}$\uparrow$3.28} & 79.133~{\scriptsize\color{ForestGreen}$\uparrow$3.73} & 81.907~{\scriptsize\color{ForestGreen}$\uparrow$2.84} & 68.833~{\scriptsize\color{ForestGreen}$\uparrow$1.83} & 75.205~{\scriptsize\color{ForestGreen}$\uparrow$1.55} & 78.011~{\scriptsize\color{ForestGreen}$\uparrow$3.17} & 81.422~{\scriptsize\color{ForestGreen}$\uparrow$2.56} \\
& AvisC~\emph{\scriptsize  ACL'25} & 79.100~{\scriptsize\color{red}$\downarrow$3.03} & 82.056~{\scriptsize\color{red}$\downarrow$1.82} & 71.800~{\scriptsize\color{red}$\downarrow$3.60} & 77.200~{\scriptsize\color{red}$\downarrow$1.87} & 64.400~{\scriptsize\color{red}$\downarrow$2.60} & 73.000~{\scriptsize\color{red}$\downarrow$0.66} & 71.767~{\scriptsize\color{red}$\downarrow$3.08} & 77.419~{\scriptsize\color{red}$\downarrow$1.45} \\
& MoD~\emph{\scriptsize  ACL'25} & 85.567~{\scriptsize\color{ForestGreen}$\uparrow$3.43} & 86.826~{\scriptsize\color{ForestGreen}$\uparrow$2.95} & 78.400~{\scriptsize\color{ForestGreen}$\uparrow$3.00} & 81.379~{\scriptsize\color{ForestGreen}$\uparrow$2.31} & 68.400~{\scriptsize\color{ForestGreen}$\uparrow$1.40} & 75.092~{\scriptsize\color{ForestGreen}$\uparrow$1.44} & 77.456~{\scriptsize\color{ForestGreen}$\uparrow$2.61} & 81.099~{\scriptsize\color{ForestGreen}$\uparrow$2.23} \\
\rowcolor{gray!20} 
\cellcolor{white}
& \textbf{SDPR~\emph{\scriptsize  ours}} & \textbf{87.333}~\textbf{\scriptsize\color{ForestGreen}$\uparrow$5.20} & \textbf{88.147}~\textbf{\scriptsize\color{ForestGreen}$\uparrow$4.27} & \textbf{79.666}~\textbf{\scriptsize\color{ForestGreen}$\uparrow$4.27} & \textbf{82.246}~\textbf{\scriptsize\color{ForestGreen}$\uparrow$3.18} & \textbf{69.400}~\textbf{\scriptsize\color{ForestGreen}$\uparrow$2.40} & \textbf{75.375}~\textbf{\scriptsize\color{ForestGreen}$\uparrow$1.72} & \textbf{78.799}~\textbf{\scriptsize\color{ForestGreen}$\uparrow$3.96} & \textbf{81.923}~\textbf{\scriptsize\color{ForestGreen}$\uparrow$3.06} \\ 
\hline

\multirow{9}{*}{\rotatebox{90}{GQA}}     
& Vanilla & 82.100 {\scriptsize\color{ForestGreen}$\uparrow$0.00} & 83.917 {\scriptsize\color{ForestGreen}$\uparrow$0.00} & 72.000 {\scriptsize\color{ForestGreen}$\uparrow$0.00} & 76.923 {\scriptsize\color{ForestGreen}$\uparrow$0.00} & 67.933 {\scriptsize\color{ForestGreen}$\uparrow$0.00} & 74.360 {\scriptsize\color{ForestGreen}$\uparrow$0.00} & 74.011 {\scriptsize\color{ForestGreen}$\uparrow$0.00} & 78.400 {\scriptsize\color{ForestGreen}$\uparrow$0.00} \\
& VCD~\emph{\scriptsize  CVPR'24} & 81.966~{\scriptsize\color{red}$\downarrow$0.13} & 84.008~{\scriptsize\color{ForestGreen}$\uparrow$0.09} & 70.733~{\scriptsize\color{red}$\downarrow$1.27} & 76.397~{\scriptsize\color{red}$\downarrow$0.53} & 68.233~{\scriptsize\color{ForestGreen}$\uparrow$0.30} & 74.901~{\scriptsize\color{ForestGreen}$\uparrow$0.54} & 73.644~{\scriptsize\color{red}$\downarrow$0.37} & 78.435~{\scriptsize\color{ForestGreen}$\uparrow$0.04} \\
& M3ID~\emph{\scriptsize  CVPR'24} & 83.100~{\scriptsize\color{ForestGreen}$\uparrow$1.00} & 84.815~{\scriptsize\color{ForestGreen}$\uparrow$0.90} & 72.166~{\scriptsize\color{ForestGreen}$\uparrow$0.17} & 77.229~{\scriptsize\color{ForestGreen}$\uparrow$0.31} & 68.533~{\scriptsize\color{ForestGreen}$\uparrow$0.60} & 75.039~{\scriptsize\color{ForestGreen}$\uparrow$0.68} & 74.600~{\scriptsize\color{ForestGreen}$\uparrow$0.59} & 79.028~{\scriptsize\color{ForestGreen}$\uparrow$0.63} \\
& VAR~\emph{\scriptsize  ICLR'25} & 82.033~{\scriptsize\color{red}$\downarrow$0.07} & 83.857~{\scriptsize\color{red}$\downarrow$0.06} & 72.066~{\scriptsize\color{ForestGreen}$\uparrow$0.07} & 76.965~{\scriptsize\color{ForestGreen}$\uparrow$0.04} & 67.966~{\scriptsize\color{ForestGreen}$\uparrow$0.03} & 74.366~{\scriptsize\color{ForestGreen}$\uparrow$0.01} & 74.022~{\scriptsize\color{ForestGreen}$\uparrow$0.01} & 78.400~{\scriptsize\color{ForestGreen}$\uparrow$0.00} \\
& VAF~\emph{\scriptsize  CVPR'25} & 79.666~{\scriptsize\color{red}$\downarrow$2.43} & 82.564~{\scriptsize\color{red}$\downarrow$1.35} & 70.000~{\scriptsize\color{red}$\downarrow$2.00} & 75.974~{\scriptsize\color{red}$\downarrow$0.95} & 66.433~{\scriptsize\color{red}$\downarrow$1.50} & 73.823~{\scriptsize\color{red}$\downarrow$0.54} & 72.033~{\scriptsize\color{red}$\downarrow$1.98} & 77.454~{\scriptsize\color{red}$\downarrow$0.95} \\
& ONLY~\emph{\scriptsize  ICCV'25} & 86.333~{\scriptsize\color{ForestGreen}$\uparrow$4.23} & 87.785~{\scriptsize\color{ForestGreen}$\uparrow$3.87} & 74.033~{\scriptsize\color{ForestGreen}$\uparrow$2.03} & 78.721~{\scriptsize\color{ForestGreen}$\uparrow$1.80} & 69.266~{\scriptsize\color{ForestGreen}$\uparrow$1.33} & 75.574~{\scriptsize\color{ForestGreen}$\uparrow$1.21} & 76.544~{\scriptsize\color{ForestGreen}$\uparrow$2.53} & 80.693~{\scriptsize\color{ForestGreen}$\uparrow$2.29} \\
& AvisC~\emph{\scriptsize  ACL'25} & 79.000~{\scriptsize\color{red}$\downarrow$3.10} & 82.200~{\scriptsize\color{red}$\downarrow$1.72} & 67.400~{\scriptsize\color{red}$\downarrow$4.60} & 74.800~{\scriptsize\color{red}$\downarrow$2.12} & 64.100~{\scriptsize\color{red}$\downarrow$3.83} & 72.900~{\scriptsize\color{red}$\downarrow$1.46} & 70.167~{\scriptsize\color{red}$\downarrow$3.84} & 76.633~{\scriptsize\color{red}$\downarrow$1.77} \\
& MoD~\emph{\scriptsize  ACL'25} & 85.766~{\scriptsize\color{ForestGreen}$\uparrow$3.67} & 87.087~{\scriptsize\color{ForestGreen}$\uparrow$3.17} & 73.533~{\scriptsize\color{ForestGreen}$\uparrow$1.53} & 78.412~{\scriptsize\color{ForestGreen}$\uparrow$1.49} & 68.166~{\scriptsize\color{ForestGreen}$\uparrow$0.23} & 75.162~{\scriptsize\color{ForestGreen}$\uparrow$0.80} & 75.822~{\scriptsize\color{ForestGreen}$\uparrow$1.81} & 80.220~{\scriptsize\color{ForestGreen}$\uparrow$1.82} \\
\rowcolor{gray!20} 
\cellcolor{white}
& \textbf{SDPR~\emph{\scriptsize  ours}} & \textbf{87.166}~\textbf{\scriptsize\color{ForestGreen}$\uparrow$5.07} & \textbf{88.098}~\textbf{\scriptsize\color{ForestGreen}$\uparrow$4.18} & \textbf{74.233}~\textbf{\scriptsize\color{ForestGreen}$\uparrow$2.23} & \textbf{78.886}~\textbf{\scriptsize\color{ForestGreen}$\uparrow$1.96} & \textbf{69.366}~\textbf{\scriptsize\color{ForestGreen}$\uparrow$1.43} & \textbf{75.617}~\textbf{\scriptsize\color{ForestGreen}$\uparrow$1.26} & \textbf{76.922}~\textbf{\scriptsize\color{ForestGreen}$\uparrow$2.91} & \textbf{80.867}~\textbf{\scriptsize\color{ForestGreen}$\uparrow$2.47} \\ 
\Xhline{0.8pt} 
\end{tabular}
}
\end{table*}
\begin{table*}[h!]
\vspace{-5pt}
\begin{minipage}{0.48\textwidth}{
    \centering
    \caption{\underline{MME} performance comparison. Models are evaluated across three different backbones (LLaVA, Qwen, and InternVL) stacked vertically.}
    \vspace{-5pt}
    \resizebox{\textwidth}{!}{
    \label{MME_table}
\begin{tabular}{c|l|ccccc} 
\Xhline{0.8pt}
\multicolumn{2}{c|}{\emph{\textbf{MME}}} & Coarse.$\uparrow$& Fine.$\uparrow$& OCR$\uparrow$& Reason.$\uparrow$& Total$\uparrow$\\ 
\hline
\multirow{9}{*}{\rotatebox{90}{LLaVA 1.5}} 
& Vanilla & 590.0~{\scriptsize\color{ForestGreen}$\uparrow$0.0} & 607.7~{\scriptsize\color{ForestGreen}$\uparrow$0.0} & 90.0~{\scriptsize\color{ForestGreen}$\uparrow$0.0} & 360.4~{\scriptsize\color{ForestGreen}$\uparrow$0.0} & 1648.0~{\scriptsize\color{ForestGreen}$\uparrow$0.0} \\
& VCD & 608.3~{\scriptsize\color{ForestGreen}$\uparrow$18.} & 639.5~{\scriptsize\color{ForestGreen}$\uparrow$32.} & 97.5~{\scriptsize\color{ForestGreen}$\uparrow$7.5} & 327.9~{\scriptsize\color{red}$\downarrow$33.} & 1673.2~{\scriptsize\color{ForestGreen}$\uparrow$25.} \\
& M3ID & 605.0~{\scriptsize\color{ForestGreen}$\uparrow$15.} & 622.8~{\scriptsize\color{ForestGreen}$\uparrow$15.} & 90.0~{\scriptsize\color{ForestGreen}$\uparrow$0.0} & 362.5~{\scriptsize\color{ForestGreen}$\uparrow$2.1} & 1680.3~{\scriptsize\color{ForestGreen}$\uparrow$32.} \\
& VAR & 551.7~{\scriptsize\color{red}$\downarrow$38.} & 601.1~{\scriptsize\color{red}$\downarrow$6.6} & 105.0~{\scriptsize\color{ForestGreen}$\uparrow$15.} & 366.8~{\scriptsize\color{ForestGreen}$\uparrow$6.4} & 1624.6~{\scriptsize\color{red}$\downarrow$23.} \\
& VAF & 596.7~{\scriptsize\color{ForestGreen}$\uparrow$6.7} & 609.1~{\scriptsize\color{ForestGreen}$\uparrow$1.4} & 112.5~{\scriptsize\color{ForestGreen}$\uparrow$23.} & 357.5~{\scriptsize\color{red}$\downarrow$2.9} & 1675.8~{\scriptsize\color{ForestGreen}$\uparrow$28.} \\
& ONLY  & 603.3~{\scriptsize\color{ForestGreen}$\uparrow$13.} & \textbf{703.0}~{\scriptsize\color{ForestGreen}$\uparrow$95.} & 120.0~{\scriptsize\color{ForestGreen}$\uparrow$30.} & 362.9~{\scriptsize\color{ForestGreen}$\uparrow$2.5} & 1789.2~{\scriptsize\color{ForestGreen}$\uparrow$141.} \\
& AvisC & 616.7~{\scriptsize\color{ForestGreen}$\uparrow$27.} & 689.6~{\scriptsize\color{ForestGreen}$\uparrow$82.} & 120.0~{\scriptsize\color{ForestGreen}$\uparrow$30.} & 354.3~{\scriptsize\color{red}$\downarrow$6.1} & 1780.5~{\scriptsize\color{ForestGreen}$\uparrow$133.} \\
& MoD & 615.0~{\scriptsize\color{ForestGreen}$\uparrow$25.} & 694.5~{\scriptsize\color{ForestGreen}$\uparrow$87.} & 115.0~{\scriptsize\color{ForestGreen}$\uparrow$25.} & 319.6~{\scriptsize\color{red}$\downarrow$41.} & 1744.1~{\scriptsize\color{ForestGreen}$\uparrow$96.} \\
\rowcolor{gray!20}\cellcolor{white}
& \textbf{SDPR} & \textbf{626.7}~\textbf{{\scriptsize\color{ForestGreen}$\uparrow$37.}} & {702.1}~\textbf{{\scriptsize\color{ForestGreen}$\uparrow$94.}} & \textbf{125.0}~\textbf{{\scriptsize\color{ForestGreen}$\uparrow$35.}} & \textbf{402.1}~\textbf{{\scriptsize\color{ForestGreen}$\uparrow$42.}} & \textbf{1855.9}~\textbf{{\scriptsize\color{ForestGreen}$\uparrow$208.}} \\
\hline
\multirow{9}{*}{\rotatebox{90}{Qwen2.5-VL}} 
& Vanilla & 621.7~{\scriptsize\color{ForestGreen}$\uparrow$0.0} & 746.6~{\scriptsize\color{ForestGreen}$\uparrow$0.0} & 110.0~{\scriptsize\color{ForestGreen}$\uparrow$0.0} & 356.4~{\scriptsize\color{ForestGreen}$\uparrow$0.0} & 1827.2~{\scriptsize\color{ForestGreen}$\uparrow$0.0} \\
& VCD & 628.3~{\scriptsize\color{ForestGreen}$\uparrow$6.6} & 792.2~{\scriptsize\color{ForestGreen}$\uparrow$46.} & 80.0~{\scriptsize\color{red}$\downarrow$30.} & 365.7~{\scriptsize\color{ForestGreen}$\uparrow$9.3} & 1881.3~{\scriptsize\color{ForestGreen}$\uparrow$54.} \\
& M3ID & 638.3~{\scriptsize\color{ForestGreen}$\uparrow$17.} & 777.1~{\scriptsize\color{ForestGreen}$\uparrow$31.} & 110.0~{\scriptsize\color{ForestGreen}$\uparrow$0.0} & 344.3~{\scriptsize\color{red}$\downarrow$12.} & 1862.2~{\scriptsize\color{ForestGreen}$\uparrow$35.} \\
& VAR & 620.0~{\scriptsize\color{red}$\downarrow$1.7} & 748.6~{\scriptsize\color{ForestGreen}$\uparrow$2.0} & 102.5~{\scriptsize\color{red}$\downarrow$7.5} & 294.6~{\scriptsize\color{red}$\downarrow$62.} & 1765.7~{\scriptsize\color{red}$\downarrow$62.} \\
& VAF & 653.3~{\scriptsize\color{ForestGreen}$\uparrow$32.} & 742.3~{\scriptsize\color{red}$\downarrow$4.3} & 110.0~{\scriptsize\color{ForestGreen}$\uparrow$0.0} & 281.8~{\scriptsize\color{red}$\downarrow$75.} & 1779.9~{\scriptsize\color{red}$\downarrow$47.} \\
& ONLY & 651.7~{\scriptsize\color{ForestGreen}$\uparrow$30.} & 803.6~{\scriptsize\color{ForestGreen}$\uparrow$57.} & 110.0~{\scriptsize\color{ForestGreen}$\uparrow$0.0} & 350.4~{\scriptsize\color{red}$\downarrow$6.0} & 1900.6~{\scriptsize\color{ForestGreen}$\uparrow$73.} \\
& AvisC & 651.7~{\scriptsize\color{ForestGreen}$\uparrow$30.} & \textbf{806.7}~{\scriptsize\color{ForestGreen}$\uparrow$60.} & 80.0~{\scriptsize\color{red}$\downarrow$30.} & 338.6~{\scriptsize\color{red}$\downarrow$18.} & 1899.4~{\scriptsize\color{ForestGreen}$\uparrow$72.} \\
& MoD & 648.3~{\scriptsize\color{ForestGreen}$\uparrow$27.} & 805.8~{\scriptsize\color{ForestGreen}$\uparrow$59.} & 95.0~{\scriptsize\color{red}$\downarrow$15.} & 371.4~{\scriptsize\color{ForestGreen}$\uparrow$15.} & 1920.6~{\scriptsize\color{ForestGreen}$\uparrow$93.} \\
\rowcolor{gray!20}\cellcolor{white}
& \textbf{SDPR} & \textbf{666.7}~\textbf{{\scriptsize\color{ForestGreen}$\uparrow$45.}} & {802.9}~\textbf{{\scriptsize\color{ForestGreen}$\uparrow$56.}} & \textbf{110.0}~\textbf{{\scriptsize\color{ForestGreen}$\uparrow$0.0}} & \textbf{373.6}~\textbf{{\scriptsize\color{ForestGreen}$\uparrow$17.}} & \textbf{1938.1}~\textbf{{\scriptsize\color{ForestGreen}$\uparrow$111.}} \\
\hline
\multirow{9}{*}{\rotatebox{90}{InternVL2.5}} 
& Vanilla & 531.7~{\scriptsize\color{ForestGreen}$\uparrow$0.0} & 708.4~{\scriptsize\color{ForestGreen}$\uparrow$0.0} & 110.0~{\scriptsize\color{ForestGreen}$\uparrow$0.0} & 429.6~{\scriptsize\color{ForestGreen}$\uparrow$0.0} & 1779.7~{\scriptsize\color{ForestGreen}$\uparrow$0.0} \\
& VCD & 605.0~{\scriptsize\color{ForestGreen}$\uparrow$73.} & 721.9~{\scriptsize\color{ForestGreen}$\uparrow$14.} & 80.0~{\scriptsize\color{red}$\downarrow$30.} & 452.1~{\scriptsize\color{ForestGreen}$\uparrow$23.} & 1859.1~{\scriptsize\color{ForestGreen}$\uparrow$79.} \\
& M3ID & 616.7~{\scriptsize\color{ForestGreen}$\uparrow$85.} & 740.6~{\scriptsize\color{ForestGreen}$\uparrow$32.} & 110.0~{\scriptsize\color{ForestGreen}$\uparrow$0.0} & 450.7~{\scriptsize\color{ForestGreen}$\uparrow$21.} & 1918.0~{\scriptsize\color{ForestGreen}$\uparrow$138.} \\
& VAR & 561.7~{\scriptsize\color{ForestGreen}$\uparrow$30.} & 708.6~{\scriptsize\color{ForestGreen}$\uparrow$0.2} & 102.5~{\scriptsize\color{red}$\downarrow$7.5} & 430.4~{\scriptsize\color{ForestGreen}$\uparrow$0.8} & 1803.1~{\scriptsize\color{ForestGreen}$\uparrow$23.} \\
& VAF & 603.3~{\scriptsize\color{ForestGreen}$\uparrow$72.} & 705.5~{\scriptsize\color{red}$\downarrow$2.9} & 110.0~{\scriptsize\color{ForestGreen}$\uparrow$0.0} & 448.3~{\scriptsize\color{ForestGreen}$\uparrow$19.} & 1867.2~{\scriptsize\color{ForestGreen}$\uparrow$88.} \\
& ONLY & 626.7~{\scriptsize\color{ForestGreen}$\uparrow$95.} & 761.9~{\scriptsize\color{ForestGreen}$\uparrow$54.} & 110.0~{\scriptsize\color{ForestGreen}$\uparrow$0.0} & 446.4~{\scriptsize\color{ForestGreen}$\uparrow$17.} & 1945.0~{\scriptsize\color{ForestGreen}$\uparrow$165.} \\
& AvisC & 641.7~{\scriptsize\color{ForestGreen}$\uparrow$110.} & 764.1~{\scriptsize\color{ForestGreen}$\uparrow$56.} & 80.0~{\scriptsize\color{red}$\downarrow$30.} & \textbf{531.1}~{\scriptsize\color{ForestGreen}$\uparrow$102.} & 2016.9~{\scriptsize\color{ForestGreen}$\uparrow$237.} \\
& MoD & 643.3~{\scriptsize\color{ForestGreen}$\uparrow$112.} & \textbf{768.0}~{\scriptsize\color{ForestGreen}$\uparrow$60.} & 95.0~{\scriptsize\color{red}$\downarrow$15.} & 516.1~{\scriptsize\color{ForestGreen}$\uparrow$87.} & 2022.4~{\scriptsize\color{ForestGreen}$\uparrow$243.} \\
\rowcolor{gray!20}\cellcolor{white}
& \textbf{SDPR} & \textbf{668.3}~\textbf{{\scriptsize\color{ForestGreen}$\uparrow$137.}} & {759.1}~{{\scriptsize\color{ForestGreen}$\uparrow$51.}} & \textbf{110.0}~\textbf{{\scriptsize\color{ForestGreen}$\uparrow$0.0}} & {528.2}~\textbf{{\scriptsize\color{ForestGreen}$\uparrow$99.}} & \textbf{2065.6}~\textbf{{\scriptsize\color{ForestGreen}$\uparrow$286.}} \\
\Xhline{0.8pt}
\end{tabular}
    }}
    \end{minipage}
    \hfill
\begin{minipage}{0.49\textwidth}
    \centering
    \vspace{3pt}
    \caption{\underline{CHAIR} performance comparison. Results for 32/64-token settings across three stacked backbones.}
    \vspace{-8pt}
\resizebox{\textwidth}{!}{
\label{chair_table}
\begin{tabular}{c|l|ccc|ccc} 
\Xhline{0.8pt}
\multicolumn{2}{c|}{\multirow{2}{*}{\normalsize \emph{\textbf{CHAIR}}}} & \multicolumn{3}{c|}{{32 Tokens}} & \multicolumn{3}{c}{{64 Tokens}} \\
\multicolumn{2}{c|}{} & $C_{S}$ $\downarrow$ & $C_{I}$ $\downarrow$ & Pre $\uparrow$ & $C_{S}$ $\downarrow$ & $C_{I}$ $\downarrow$ & Pre $\uparrow$ \\ 
\hline
\multirow{9}{*}{\rotatebox{90}{LLaVA 1.5}} 
& Vanilla & 10.8~{\scriptsize\color{ForestGreen}$\downarrow$0.0} & 6.0~{\scriptsize\color{ForestGreen}$\downarrow$0.0} & 87.8~{\scriptsize\color{ForestGreen}$\uparrow$0.0} & 21.2~{\scriptsize\color{ForestGreen}$\downarrow$0.0} & 7.4~{\scriptsize\color{ForestGreen}$\downarrow$0.0} & 89.5~{\scriptsize\color{ForestGreen}$\uparrow$0.0} \\
& VCD & 10.4~{\scriptsize\color{ForestGreen}$\downarrow$0.4} & 4.8~{\scriptsize\color{ForestGreen}$\downarrow$1.2} & 91.5~{\scriptsize\color{ForestGreen}$\uparrow$3.7} & 21.4~{\scriptsize\color{red}$\uparrow$0.2} & 7.5~{\scriptsize\color{red}$\uparrow$0.1} & 90.5~{\scriptsize\color{ForestGreen}$\uparrow$1.0} \\
& M3ID & 9.0~{\scriptsize\color{ForestGreen}$\downarrow$1.8} & 4.0~{\scriptsize\color{ForestGreen}$\downarrow$2.0} & 92.2~{\scriptsize\color{ForestGreen}$\uparrow$4.4} & 22.4~{\scriptsize\color{red}$\uparrow$1.2} & 6.9~{\scriptsize\color{ForestGreen}$\downarrow$0.5} & 90.2~{\scriptsize\color{ForestGreen}$\uparrow$0.7} \\
& VAF & 9.8~{\scriptsize\color{ForestGreen}$\downarrow$1.0} & 5.5~{\scriptsize\color{ForestGreen}$\downarrow$0.5} & 88.3~{\scriptsize\color{ForestGreen}$\uparrow$0.5} & 23.6~{\scriptsize\color{red}$\uparrow$2.4} & 8.9~{\scriptsize\color{red}$\uparrow$1.5} & 87.9~{\scriptsize\color{red}$\downarrow$1.6} \\
& VAR & 10.6~{\scriptsize\color{ForestGreen}$\downarrow$0.2} & 5.6~{\scriptsize\color{ForestGreen}$\downarrow$0.4} & 87.8~{\scriptsize\color{ForestGreen}$\uparrow$0.0} & 22.2~{\scriptsize\color{red}$\uparrow$1.0} & 7.6~{\scriptsize\color{red}$\uparrow$0.2} & 87.3~{\scriptsize\color{red}$\downarrow$2.2} \\
& ONLY & 8.4~{\scriptsize\color{ForestGreen}$\downarrow$2.4} & 4.2~{\scriptsize\color{ForestGreen}$\downarrow$1.8} & 92.9~{\scriptsize\color{ForestGreen}$\uparrow$5.1} & 17.2~{\scriptsize\color{ForestGreen}$\downarrow$4.0} & 5.3~{\scriptsize\color{ForestGreen}$\downarrow$2.1} & 91.3~{\scriptsize\color{ForestGreen}$\uparrow$1.8} \\
& AvisC & 7.5~{\scriptsize\color{ForestGreen}$\downarrow$3.3} & 3.6~{\scriptsize\color{ForestGreen}$\downarrow$2.4} & 93.6~{\scriptsize\color{ForestGreen}$\uparrow$5.8} & 20.6~{\scriptsize\color{ForestGreen}$\downarrow$0.6} & 6.5~{\scriptsize\color{ForestGreen}$\downarrow$0.9} & 91.2~{\scriptsize\color{ForestGreen}$\uparrow$1.7} \\
& MoD & 7.4~{\scriptsize\color{ForestGreen}$\downarrow$3.4} & 3.4~{\scriptsize\color{ForestGreen}$\downarrow$2.6} & 94.3~{\scriptsize\color{ForestGreen}$\uparrow$6.5} & 20.6~{\scriptsize\color{ForestGreen}$\downarrow$0.6} & 6.4~{\scriptsize\color{ForestGreen}$\downarrow$1.0} & 91.2~{\scriptsize\color{ForestGreen}$\uparrow$1.7} \\
\rowcolor{gray!20}\cellcolor{white}
& \textbf{SDPR} & \textbf{7.2}~\textbf{{\scriptsize\color{ForestGreen}$\downarrow$3.6}} & \textbf{3.4}~\textbf{{\scriptsize\color{ForestGreen}$\downarrow$2.6}} & \textbf{93.1}~\textbf{{\scriptsize\color{ForestGreen}$\uparrow$5.3}} & \textbf{15.2}~\textbf{{\scriptsize\color{ForestGreen}$\downarrow$6.0}} & \textbf{4.7}~\textbf{{\scriptsize\color{ForestGreen}$\downarrow$2.7}} & \textbf{92.5}~\textbf{{\scriptsize\color{ForestGreen}$\uparrow$3.0}} \\
\hline
\multirow{9}{*}{\rotatebox{90}{Qwen2.5-VL}} 
& Vanilla & 14.4~{\scriptsize\color{ForestGreen}$\downarrow$0.0} & 5.8~{\scriptsize\color{ForestGreen}$\downarrow$0.0} & 92.3~{\scriptsize\color{ForestGreen}$\uparrow$0.0} & 31.6~{\scriptsize\color{ForestGreen}$\downarrow$0.0} & 10.3~{\scriptsize\color{ForestGreen}$\downarrow$0.0} & 86.5~{\scriptsize\color{ForestGreen}$\uparrow$0.0} \\
& VCD & 10.6~{\scriptsize\color{ForestGreen}$\downarrow$3.8} & 5.8~{\scriptsize\color{ForestGreen}$\downarrow$0.0} & 90.6~{\scriptsize\color{red}$\downarrow$1.7} & 33.0~{\scriptsize\color{red}$\uparrow$1.4} & 10.4~{\scriptsize\color{red}$\uparrow$0.1} & 84.3 {\scriptsize\color{red}$\downarrow$2.2} \\
& M3ID & 8.0~{\scriptsize\color{ForestGreen}$\downarrow$6.4} & 5.1~{\scriptsize\color{ForestGreen}$\downarrow$0.7} & 92.2~{\scriptsize\color{red}$\downarrow$0.1} & 33.2~{\scriptsize\color{red}$\uparrow$1.6} & 10.6~{\scriptsize\color{red}$\uparrow$0.3} & 84.2~{\scriptsize\color{red}$\downarrow$2.3}  \\
& VAF & 16.4~{\scriptsize\color{red}$\uparrow$2.0} & 6.5~{\scriptsize\color{red}$\uparrow$0.7} & 91.0~{\scriptsize\color{red}$\downarrow$1.3} & 30.4~{\scriptsize\color{ForestGreen}$\downarrow$1.2} & 10.4~{\scriptsize\color{red}$\uparrow$0.1} & 86.7~{\scriptsize\color{ForestGreen}$\uparrow$0.2} \\
& VAR & 15.4~{\scriptsize\color{red}$\uparrow$1.0} & 6.0~{\scriptsize\color{red}$\uparrow$0.2} & 92.0~{\scriptsize\color{red}$\downarrow$0.3} & 32.6~{\scriptsize\color{red}$\uparrow$1.0} & 10.6~{\scriptsize\color{red}$\uparrow$0.3} & 85.8~{\scriptsize\color{red}$\downarrow$0.7} \\
& ONLY & 8.0~{\scriptsize\color{ForestGreen}$\downarrow$6.4} & 4.2~{\scriptsize\color{ForestGreen}$\downarrow$1.6} & 93.0~{\scriptsize\color{ForestGreen}$\uparrow$0.7} & 27.3~{\scriptsize\color{ForestGreen}$\downarrow$4.3} & 8.4~{\scriptsize\color{ForestGreen}$\downarrow$1.9} & 89.6~{\scriptsize\color{ForestGreen}$\uparrow$3.1} \\
& AvisC & 15.2~{\scriptsize\color{red}$\uparrow$0.8} & 6.2~{\scriptsize\color{red}$\uparrow$0.4} & 92.0~{\scriptsize\color{red}$\downarrow$0.3} & 32.2~{\scriptsize\color{red}$\uparrow$0.6} & 10.0~{\scriptsize\color{ForestGreen}$\downarrow$0.3} & 85.8~{\scriptsize\color{red}$\downarrow$0.7} \\
& MoD & 14.4~{\scriptsize\color{ForestGreen}$\downarrow$0.0} & 5.8~{\scriptsize\color{ForestGreen}$\downarrow$0.0} & 92.3~{\scriptsize\color{ForestGreen}$\uparrow$0.0} & 30.4~{\scriptsize\color{ForestGreen}$\downarrow$1.2} & 10.1~{\scriptsize\color{ForestGreen}$\downarrow$0.2} & 87.0~{\scriptsize\color{ForestGreen}$\uparrow$0.5} \\
\rowcolor{gray!20}\cellcolor{white}
& \textbf{SDPR} & \textbf{7.2}~\textbf{{\scriptsize\color{ForestGreen}$\downarrow$7.2}} & \textbf{4.2}~\textbf{{\scriptsize\color{ForestGreen}$\downarrow$1.6}} & \textbf{93.1}~\textbf{{\scriptsize\color{ForestGreen}$\uparrow$0.8}} & \textbf{24.2}~\textbf{{\scriptsize\color{ForestGreen}$\downarrow$7.4}} & \textbf{7.6}~\textbf{{\scriptsize\color{ForestGreen}$\downarrow$2.7}} & \textbf{90.2}~\textbf{{\scriptsize\color{ForestGreen}$\uparrow$3.7}} \\
\hline
\multirow{9}{*}{\rotatebox{90}{InternVL2.5}} 
& Vanilla & 6.8~{\scriptsize\color{ForestGreen}$\uparrow$0.0} & 4.1~{\scriptsize\color{ForestGreen}$\uparrow$0.0} & 80.8~{\scriptsize\color{ForestGreen}$\uparrow$0.0} & 15.4~{\scriptsize\color{ForestGreen}$\uparrow$0.0} & 6.8~{\scriptsize\color{ForestGreen}$\uparrow$0.0} & 82.1~{\scriptsize\color{ForestGreen}$\uparrow$0.0} \\
& VCD & 6.2~{\scriptsize\color{ForestGreen}$\downarrow$0.6} & 4.2~{\scriptsize\color{red}$\uparrow$0.1} & 83.8~{\scriptsize\color{ForestGreen}$\uparrow$3.0} & 16.4~{\scriptsize\color{red}$\uparrow$1.0} & 6.0~{\scriptsize\color{ForestGreen}$\downarrow$0.8} & 86.4~{\scriptsize\color{ForestGreen}$\uparrow$4.3} \\
& M3ID & 6.2~{\scriptsize\color{ForestGreen}$\downarrow$0.6} & 3.9~{\scriptsize\color{ForestGreen}$\downarrow$0.2} & 80.8~{\scriptsize\color{ForestGreen}$\uparrow$0.0} & 14.8~{\scriptsize\color{ForestGreen}$\downarrow$0.6} & 5.6~{\scriptsize\color{ForestGreen}$\downarrow$1.2} & \textbf{87.0}~{\scriptsize\color{ForestGreen}$\uparrow$4.9} \\
& VAF & 6.4~{\scriptsize\color{ForestGreen}$\downarrow$0.4} & 4.4~{\scriptsize\color{red}$\uparrow$0.3} & 77.2~{\scriptsize\color{red}$\downarrow$3.6} & 15.1~{\scriptsize\color{ForestGreen}$\downarrow$0.3} & 7.1~{\scriptsize\color{red}$\uparrow$0.3} & 82.1~{\scriptsize\color{ForestGreen}$\uparrow$0.0} \\
& VAR & 7.8~{\scriptsize\color{red}$\uparrow$1.0} & 4.5~{\scriptsize\color{red}$\uparrow$0.4} & \textbf{83.8}~{\scriptsize\color{ForestGreen}$\uparrow$3.0} & 20.2~{\scriptsize\color{red}$\uparrow$4.8} & 8.6~{\scriptsize\color{red}$\uparrow$1.8} & 84.2~{\scriptsize\color{ForestGreen}$\uparrow$2.1} \\
& ONLY & 8.2~{\scriptsize\color{red}$\uparrow$1.4} & 5.3~{\scriptsize\color{red}$\uparrow$1.2} & 81.3~{\scriptsize\color{ForestGreen}$\uparrow$0.5} & 18.6~{\scriptsize\color{red}$\uparrow$3.2} & 7.1~{\scriptsize\color{red}$\uparrow$0.3} & 85.4~{\scriptsize\color{ForestGreen}$\uparrow$3.3} \\
& AvisC & 6.2~{\scriptsize\color{ForestGreen}$\downarrow$0.6} & 3.8~{\scriptsize\color{ForestGreen}$\downarrow$0.3} & 81.6~{\scriptsize\color{ForestGreen}$\uparrow$0.8} & 16.1~{\scriptsize\color{red}$\uparrow$0.7} & 7.2~{\scriptsize\color{red}$\uparrow$0.4} & 81.0~{\scriptsize\color{red}$\downarrow$1.1} \\
& MoD & 6.3~{\scriptsize\color{ForestGreen}$\downarrow$0.5} & 3.8~{\scriptsize\color{ForestGreen}$\downarrow$0.3} & 81.8~{\scriptsize\color{ForestGreen}$\uparrow$1.0} & 15.4~{\scriptsize\color{ForestGreen}$\uparrow$0.0} & 6.8~{\scriptsize\color{ForestGreen}$\uparrow$0.0} & 82.1~{\scriptsize\color{ForestGreen}$\uparrow$0.0} \\
\rowcolor{gray!20}\cellcolor{white}
& \textbf{SDPR} & \textbf{6.0}~\textbf{{\scriptsize\color{ForestGreen}$\downarrow$0.8}} & \textbf{3.6}~\textbf{{\scriptsize\color{ForestGreen}$\downarrow$0.5}} & {82.2}~\textbf{{\scriptsize\color{ForestGreen}$\uparrow$1.4}} & \textbf{14.6}~\textbf{{\scriptsize\color{ForestGreen}$\downarrow$0.8}} & \textbf{5.3}~\textbf{{\scriptsize\color{ForestGreen}$\downarrow$1.5}} & {85.8}~\textbf{{\scriptsize\color{ForestGreen}$\uparrow$3.7}} \\
\Xhline{0.8pt}
\end{tabular}
    
    }
    \end{minipage}
\end{table*}

\begin{table}[h!]
\centering
\vspace{-3pt}
\caption{Ablation study on primary modules.}
\label{ablation_study_1}
\vspace{-10pt}
\resizebox{\linewidth}{!}{
\begin{tabular}{ccc|ccccc}
\Xhline{0.8pt}
\emph{SDAR} & \emph{SDCA} & \emph{PCD} & {MME} $\uparrow$ & {Coarse.} $\uparrow$ & {$C_s$} $\downarrow$ & {$C_I$} $\downarrow$ & {Pre} $\uparrow$ \\
\hline
\textcolor{red}{\ding{55}} & \textcolor{red}{\ding{55}} & \textcolor{red}{\ding{55}} & 1648.0~{\scriptsize\color{ForestGreen} $\uparrow$0.00} & 590.0~{\scriptsize\color{ForestGreen}$\uparrow$0.0} & 10.8~{\scriptsize\color{ForestGreen}$\uparrow$0.0} & 6.0~{\scriptsize\color{ForestGreen}$\uparrow$0.0} & 87.8~{\scriptsize\color{ForestGreen}$\uparrow$0.0} \\

\textcolor{green}{\ding{51}} & \textcolor{red}{\ding{55}} & \textcolor{red}{\ding{55}} & 1688.0~{\scriptsize\color{ForestGreen}$\uparrow$40.0} & 606.7~{\scriptsize\color{ForestGreen}$\uparrow$17.} & 10.4~{\scriptsize\color{ForestGreen}$\downarrow$0.4} & 5.6~{\scriptsize\color{ForestGreen}$\downarrow$0.4} & 88.1~{\scriptsize\color{ForestGreen}$\uparrow$0.3} \\

\textcolor{red}{\ding{55}} & \textcolor{green}{\ding{51}} & \textcolor{red}{\ding{55}} & 1680.2~{\scriptsize\color{ForestGreen}$\uparrow$32.2} & 603.0~{\scriptsize\color{ForestGreen}$\uparrow$13.} & 10.2~{\scriptsize\color{ForestGreen}$\downarrow$0.6} & 5.3~{\scriptsize\color{ForestGreen}$\downarrow$0.7} & 88.4~{\scriptsize\color{ForestGreen}$\uparrow$0.6} \\ 

\textcolor{green}{\ding{51}} & \textcolor{green}{\ding{51}} & \textcolor{red}{\ding{55}} & 1692.6~{\scriptsize\color{ForestGreen}$\uparrow$44.6} & 619.2~{\scriptsize\color{ForestGreen}$\uparrow$29.} & 10.2~{\scriptsize\color{ForestGreen}$\downarrow$0.6} & 5.0~{\scriptsize\color{ForestGreen}$\downarrow$1.0} & 88.4~{\scriptsize\color{ForestGreen}$\uparrow$0.6} \\

\textcolor{red}{\ding{55}} & \textcolor{red}{\ding{55}} & \textcolor{green}{\ding{51}} & 1779.0~{\scriptsize\color{ForestGreen}$\uparrow$131.} & 602.0~{\scriptsize\color{ForestGreen}$\uparrow$12.} & 8.4~{\scriptsize\color{ForestGreen}$\downarrow$2.4} & 4.2~{\scriptsize\color{ForestGreen}$\downarrow$1.8} & 92.5~{\scriptsize\color{ForestGreen}$\uparrow$4.7} \\

\textcolor{red}{\ding{55}} & \textcolor{green}{\ding{51}} & \textcolor{green}{\ding{51}} & 1811.4~{\scriptsize\color{ForestGreen}$\uparrow$163.} & 615.2~{\scriptsize\color{ForestGreen}$\uparrow$25.} & 8.2~{\scriptsize\color{ForestGreen}$\downarrow$2.6} & 4.2~{\scriptsize\color{ForestGreen}$\downarrow$1.8} & 92.9~{\scriptsize\color{ForestGreen}$\uparrow$5.1} \\

\textcolor{green}{\ding{51}} & \textcolor{red}{\ding{55}} & \textcolor{green}{\ding{51}} & 1841.2~{\scriptsize\color{ForestGreen}$\uparrow$193.} & 618.3~{\scriptsize\color{ForestGreen}$\uparrow$28.} & 8.2~{\scriptsize\color{ForestGreen}$\downarrow$2.6} & 4.2~{\scriptsize\color{ForestGreen}$\downarrow$1.8} & 92.6~{\scriptsize\color{ForestGreen}$\uparrow$4.8} \\

\rowcolor{gray!20}
\cellcolor{white}\textbf{\textcolor{green}{\ding{51}}}
& \cellcolor{white}\textbf{\textcolor{green}{\ding{51}}}
& \cellcolor{white}\textbf{\textcolor{green}{\ding{51}}}
& \textbf{1855.9~{\scriptsize\color{ForestGreen}$\uparrow$208.}}
& \textbf{626.7~{\scriptsize\color{ForestGreen}$\uparrow$37.}}
& \textbf{7.2~{\scriptsize\color{ForestGreen}$\downarrow$3.6}}
& \textbf{3.4~{\scriptsize\color{ForestGreen}$\downarrow$2.6}}
& \textbf{93.1~{\scriptsize\color{ForestGreen}$\uparrow$5.3}} \\

\Xhline{0.8pt}
\end{tabular}
}
\vspace{-12pt}
\end{table}

\subsection{Experimental Setups}

\textbf{Evaluated Models.}
We evaluate the proposed method across three representative open-source
LVLMs, \emph{i.e.}, LLaVA-1.5 \cite{llava_1.5}, Qwen2.5-VL
\cite{qwen}, and InternVL2.5 \cite{internvl}.\\
\textbf{Benchmark Datasets.}
We conduct experiments on three hallucination benchmarks,
\emph{i.e.}, POPE \cite{dataset_pope}, CHAIR \cite{dataset_chair},
and MME \cite{dataset_mme}. We further evaluate our method on general-purpose
LVLM benchmarks, including MMMU \cite{dataset_mmmu}, MMMU-Pro
\cite{dataset_mmmupro}, MME-RealWorld \cite{dataset_mmerealworld}, and MMVP
\cite{dataset_mmvp}. Results on these benchmarks are reported in the
\href{https://github.com/PengSyuChen/SDPR/blob/main/supplementary.pdf}
{supplementary material}.\\
\textbf{Baselines.}
We evaluate our approach against leading CD-based and
attention-intervention baselines: VCD \cite{vcd}, M3ID \cite{m3id},
VAR \cite{sink_iclr25_1}, VAF \cite{vaf_clearsight}, ONLY
\cite{only}, AvisC \cite{avisc}, and MoD \cite{mod}.\\
\textbf{Implementation Details.}
We adopt the default query format and KV caching for all LVLMs. For SDAR, we select 16 low-entropy heads and
redistribute attention over five candidate token positions in the middle layers of the LVLMs, where cross-modal interactions are frequent \cite{vaf_clearsight}. In SDCA, we
extract the query-relevant saliency map from the final
high-interaction layer (Layer 15). For adaptive decoding, we set
$\lambda=0.2$, $\alpha_{1}=3$, and $\alpha_{2}=1$, with $\gamma=0.2$
for LLaVA and $\gamma=0.4$ for Qwen and InternVL. 
All experiments are conducted on two NVIDIA RTX 5090 GPUs.

\subsection{Results and Discussion}
\textbf{Results on POPE.} Table~\ref{pope_llava} shows that SDPR consistently
improves accuracy and F1 scores on POPE with LLaVA-1.5. Across the COCO,
A-OKVQA, and GQA
branches, SDPR improves average accuracy by 4.33, 3.96, and 2.91
percentage points, respectively, and average F1 score by 3.68, 3.06,
and 2.47 points over the vanilla baseline. On MS-COCO, SDPR surpasses VCD, M3ID, ONLY, AvisC, and MoD in average
accuracy by 3.96, 2.99, 0.41, 6.98, and 0.69 points, respectively.
It also exceeds VAR and VAF in F1 score by 4.07 and 3.27 points. Similar trends
on Qwen2.5-VL and InternVL2.5 are reported in the supplementary
material. Overall, the consistent gains across datasets and LVLM backbones
show that SDPR robustly mitigates hallucinations despite variations in data
distribution and model architecture, confirming its broad applicability. \par
\textbf{Results on MME.}
Table~\ref{MME_table} compares SDPR with existing baselines on the
full MME benchmark across coarse-grained perception, fine-grained
perception, OCR, and reasoning tasks. We focus on coarse-grained
perception to assess object-level visual faithfulness, while reporting
the other tasks for a comprehensive evaluation of SDPR. Taking LLaVA-1.5 as an
example, SDPR improves the coarse-grained score by 36.7 points over the
vanilla baseline, 18.4 points over VCD, and 21.7 points over M3ID,
demonstrating stronger object-level visual grounding. On the
fine-grained task, SDPR exceeds VCD, M3ID, VAF, and AvisC by 62.6,
79.3, 93.0, and 12.5 points, respectively, while trailing ONLY by 0.9 points. Nevertheless, SDPR surpasses ONLY by 39.2 points on the
reasoning task and 66.7 points on the overall score. These balanced gains arise from the complementary effects of SDAR, SDCA, and PCD, which recover salient visual evidence, enhance its retrieval from the key cache, and suppress language-prior-dominated predictions, respectively.
\par
\textbf{Results on CHAIR.}
For \emph{open-ended generation} tasks, SDPR achieves lower hallucination rates.
Table~\ref{chair_table} compares SDPR with seven SOTA methods using $\text{CHAIR}_{S}$,
$\text{CHAIR}_{I}$, and precision (Pre) under maximum generation lengths of
32 and 64 tokens. For LLaVA-1.5 with a
64-token limit, SDPR reduces $\text{CHAIR}_{S}$ by 6.0 points and
$\text{CHAIR}_{I}$ by 2.7 points compared with regular decoding. It
also improves precision by 3.0 points over the vanilla baseline and by
2.0 and 1.2 points over VCD and ONLY, respectively, demonstrating its
effectiveness in long-form generation. These gains result from
coordinated interventions across perception, memory, and decoding:
SDAR recovers salient visual evidence, SDCA re-anchors query-relevant
features in the cache, and PCD suppresses unfaithful predictions dominated by
language priors. \\

\subsection{Ablation Study}
We conduct an ablation study of the \textbf{primary modules},
\emph{i.e.}, SDAR, SDCA, and PCD, using MME and CHAIR. The
\textbf{hyperparameter} studies are provided in the
\href{https://github.com/PengSyuChen/SDPR/blob/main/supplementary.pdf}
{supplementary material}. 

\textbf{Study on Primary Modules.}
Table~\ref{ablation_study_1} shows that SDAR strengthens visual perception,
raising the MME score by 40.0 points and reducing $\text{CHAIR}_{S}$ by 0.4
points. SDCA independently raises the MME score by 32.2 points and reduces
$\text{CHAIR}_{S}$ by 0.6 points, confirming the benefit of cache realignment.
Combining SDAR and SDCA further raises the coarse-grained score from
590.0 to 619.2. By suppressing predictions dominated by language priors, PCD alone
improves the MME score by 131.0 points and reduces
$\text{CHAIR}_{S}$ by 2.4 points. The complete framework achieves the best overall results, demonstrating that the three modules jointly reinforce visual grounding across perception, memory, and prediction.
\subsection{Case Study}
\begin{figure}[t!] 
 \centering 
 \vspace{-8pt} 
\includegraphics[width=0.46\textwidth]{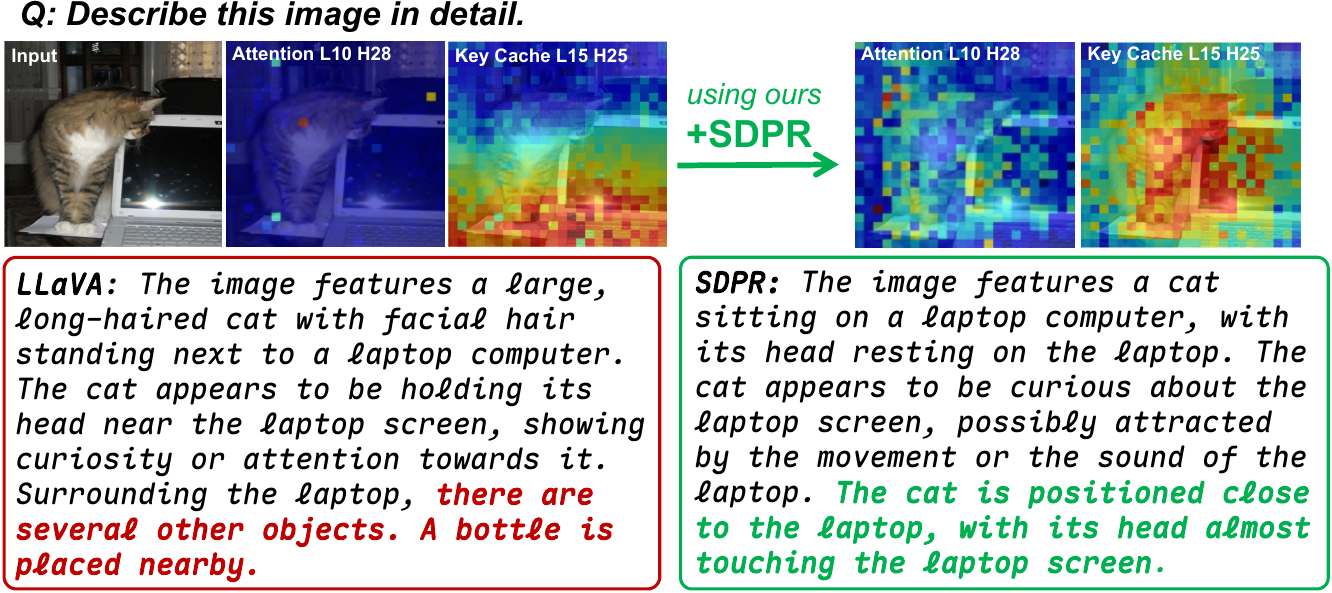}  
\vspace{-0.3cm} 
\caption{\textbf{Case Study.} Comparison of responses and internal
states between regular decoding and SDPR on LLaVA-1.5. The prompt is
``\emph{Describe this image in detail}.'' Hallucinated and faithful
content is highlighted in \textcolor{red}{\textbf{red}} and
\textcolor{ForestGreen}{\textbf{green}}, respectively.}
 \vspace{-0.4cm} 
 \label{casestudy}
\end{figure}

A qualitative case study on open-ended generation is presented in
Figure~\ref{casestudy}. LLaVA shows poor factual grounding, characterized by
pronounced object hallucinations, as exemplified by the false claim
\emph{``there are several other objects. A bottle is placed nearby.''} By
examining the internal states, we observe sink-dominated attention in MHSA and
weakened retrieval from cached visual keys, causing predictions to rely more
heavily on language priors than on visual evidence. In contrast, SDPR
redistributes hijacked attention and re-anchors query-relevant visual features
in the key cache. This perceptual realignment enables the model to capture
fine-grained interaction details, \emph{e.g.}, the cat's head \emph{``almost
touching''} the screen. By improving scene perception, SDPR produces reliable
responses and effectively mitigates hallucinations.

\section{Conclusion}
In this paper, we propose SDPR, a novel training-free intervention framework
that mitigates hallucinations in LVLMs by restoring visual awareness
throughout inference. SDPR addresses key limitations of current LVLMs,
including degraded visual grounding across perception and memory and the
resulting shift toward language-prior-dominated predictions during decoding.
SDPR holistically realigns visual awareness across perception, memory, and
decoding, supporting visually grounded reasoning without additional training.
Extensive experiments show that SDPR consistently outperforms existing
state-of-the-art methods, confirming its effectiveness in strengthening
visual grounding and mitigating hallucinations.
\par

\section{Acknowledgements}
This work was supported by grant from the Natural Science Foundation of Shaanxi Province (2024JCJCQN-66).

\bibliographystyle{unsrt}  
\bibliography{mybib}
\balance
\end{document}